\documentclass[runningheads,a4paper]{llncs}

\usepackage{amssymb}
\usepackage{epsfig}
\usepackage{graphicx}
\usepackage{verbatim}
\usepackage{hyperref}
\usepackage{amsmath}
\usepackage{subcaption}
\usepackage{cleveref}
\usepackage{setspace}
\usepackage{tikz}
\usetikzlibrary{arrows,shapes}

\usepackage{url}  
\newcommand{\keywords}[1]{\par\addvspace\baselineskip
\noindent\keywordname\enspace\ignorespaces#1}
\newcommand{\youtubeurl}{https://youtu.be/m5z0aFZgEX8}

\begin{document}

\mainmatter  % start of an individual contribution

% first the title is needed
\title{Towards Learning to Perceive \\and Reason About Liquids \vspace{-0.5cm}}

% a short form should be given in case it is too long for the running head
\titlerunning{Towards Learning to Perceive and Reason About Liquids}

% the name(s) of the author(s) follow(s) next
%
% NB: Chinese authors should write their first names(s) in front of
% their surnames. This ensures that the names appear correctly in
% the running heads and the author index.
%
\author{Connor Schenck, Dieter Fox \vspace{-0.3cm}}
\authorrunning{Towards Learning to Perceive and Reason About Liquids}
% (feature abused for this document to repeat the title also on left hand pages)

% the affiliations are given next; don't give your e-mail address
% unless you accept that it will be published
\institute{University of Washington}

%
% NB: a more complex sample for affiliations and the mapping to the
% corresponding authors can be found in the file "llncs.dem"
% (search for the string "\mainmatter" where a contribution starts).
% "llncs.dem" accompanies the document class "llncs.cls".
%

%\toctitle{Lecture Notes in Computer Science}
%\tocauthor{Authors' Instructions}
\maketitle

\vspace{-0.5cm}
\begin{abstract}
Recent advances in AI and robotics have claimed many incredible results with deep learning, yet no work to date has applied deep learning to the problem of liquid perception and reasoning. In this paper, we apply fully-convolutional deep neural networks to the tasks of detecting and tracking liquids. We evaluate three models: a single-frame network, multi-frame network, and a LSTM recurrent network. Our results show that the best liquid detection results are achieved when aggregating data over multiple frames and that the LSTM network outperforms the other two in both tasks. This suggests that LSTM-based neural networks have the potential to be a key component for enabling robots to handle liquids using robust, closed-loop controllers.
\keywords{Robot perception, deep learning, liquids, manipulation}
\end{abstract}

\section{Introduction}

To robustly handle liquids, such as pouring a certain amount of water into a
bowl, a robot must be able to perceive and reason about liquids in a way that
allows for closed-loop control. Liquids present many challenges compared to
solid objects. For example, liquids can not be interacted with directly by a
robot, instead the robot must use a tool or container; often containers
containing some amount of liquid are opaque, obstructing the robot's view of the
liquid and forcing it to remember the liquid in the container, rather than
re-perceiving it at each timestep; and finally liquids are frequently
transparent, making simply distinguishing them from the background a difficult
task. Taken together, these challenges make perceiving and manipulating liquids
highly non-trivial.

Recent advances in deep learning have enabled a leap in performance not only on visual recognition tasks, but also in areas ranging from playing Atari games \cite{guo2014} to end-to-end policy training in robotics \cite{levine2015}.  
In this paper, we investigate how deep learning techniques can be used for perceiving liquids during pouring tasks.  
We develop a method for generating large amounts of labeled pouring data for training and testing using a realistic liquid simulation and rendering engine, which we use to generate a data set with 10,122 pouring sequences, each 15 seconds long, for a total of 2,531 minutes of video or over 4.5 million labeled images. 
Using this dataset, we evaluate multiple deep learning network architectures on the tasks of detecting liquid in an image and tracking the location of liquid even when occluded. 
Our results show that deep networks able to detect and track liquid in a simulated environment with a reasonable degree of robustness. 
We also have preliminary results that show that these networks perform well in real environments.

\vspace{-0.5cm}
\section{Related Work}
\vspace{-0.3cm}

To the best of our knowledge, no prior work has investigated directly perceiving
and reasoning about liquids. Existing work relating to liquids either uses
coarse simulations that are disconnected to real liquid perception and dynamics
\cite{kunze2015,yamaguchi2015} or constrained task spaces that bypass
the need to perceive or reason directly about liquids
\cite{langsfeld2014,okada2006,tamosiunaite2011,cakmak2012,rozo2013}. 
While some of this work has dealt with pouring, none of it has attempted to directly perceive liquids from raw sensory data. 
In contrast, in this work we directly approach this problem.

Similarly, Rankin {\it et al.} \cite{rankin2010,rankin2011} investigated ways to detect pools of water from an unmanned ground vehicle navigating rough terrain. 
They detected water based on simple color features or sky reflections, and didn't reason about the dynamics of the water, instead treating it as a static obstacle. 
Griffith {\it et al.} \cite{griffith2012} learned to categorize objects based on their interactions with running water, although the robot did not detect or reason about the water itself, rather it used the water as a means to learn about the objects.
%used the auditory and proprioceptive feedback from objects interacted with in a sink environment with a running water tap in order to learn about those objects, although in this case the robot did not detect or reason about the water, rather it used the water as a means to learn about and categorize other objects. 
In contrast to \cite{griffith2012}, we use vision to directly detect the liquid itself, and unlike \cite{rankin2010,rankin2011}, we treat the liquid as dynamic and reason about it.

In order to perceive liquids at the pixel level, we make use of
fully-convolutional neural networks (FCN). FCNs have been successfully applied
to the task of image segmentation in the past
\cite{long2015,havaei2015,romera2015} and are a natural fit for pixel-wise
classification. In addition to FCNs, we utilize long short-term memory (LSTM)
\cite{hochreiter1997} recurrent cells to reason about the temporal evolution of
liquids. LSTMs are preferable over more standard recurrent networks for
long-term memory as they overcome many of the numerical issues during training
such as exploding gradients \cite{greff2015}. LSTM-based CNNs have been
successfully applied to many temporal memory tasks by previous work
\cite{junhyuk2015,romera2015}, and in fact LSTMs have even been combined with FCNs by replacing the standard fully-connected layers of their LSTMs with
$1\times1$ convolution layers \cite{romera2015}. We use a similar method in this paper.

\vspace{-0.3cm}
\section{Methodology}
\label{sec:methodology}
\vspace{-0.5cm}

In order to train neural networks to perceive and reason about liquids, we must first have labeled data to train on. 
Getting pixel-wise labels for real-world data can be difficult, so in this paper we opt to use a realistic liquid simulator. 
In this way we can acquire ground truth pixel labels while generating images that appear as realistic as possible.
We train three different types of convolutional neural networks (CNNs) on this generated data to detect and track the liquid: single-frame CNN, multi-frame CNN, and LSTM-CNN.

\vspace{-0.3cm}
\subsection{Data Generation}
\label{sec:data_gen}
\vspace{-0.3cm}

\begin{figure}[t]
    \centering
    \setlength{\fboxsep}{0pt}
    \setlength{\fboxrule}{1pt}
    \setlength{\unitlength}{1.0cm}
    \begin{picture}(6.5,6.5)
        \put(0.0,0.0){\fbox{\includegraphics[width=6.5cm]{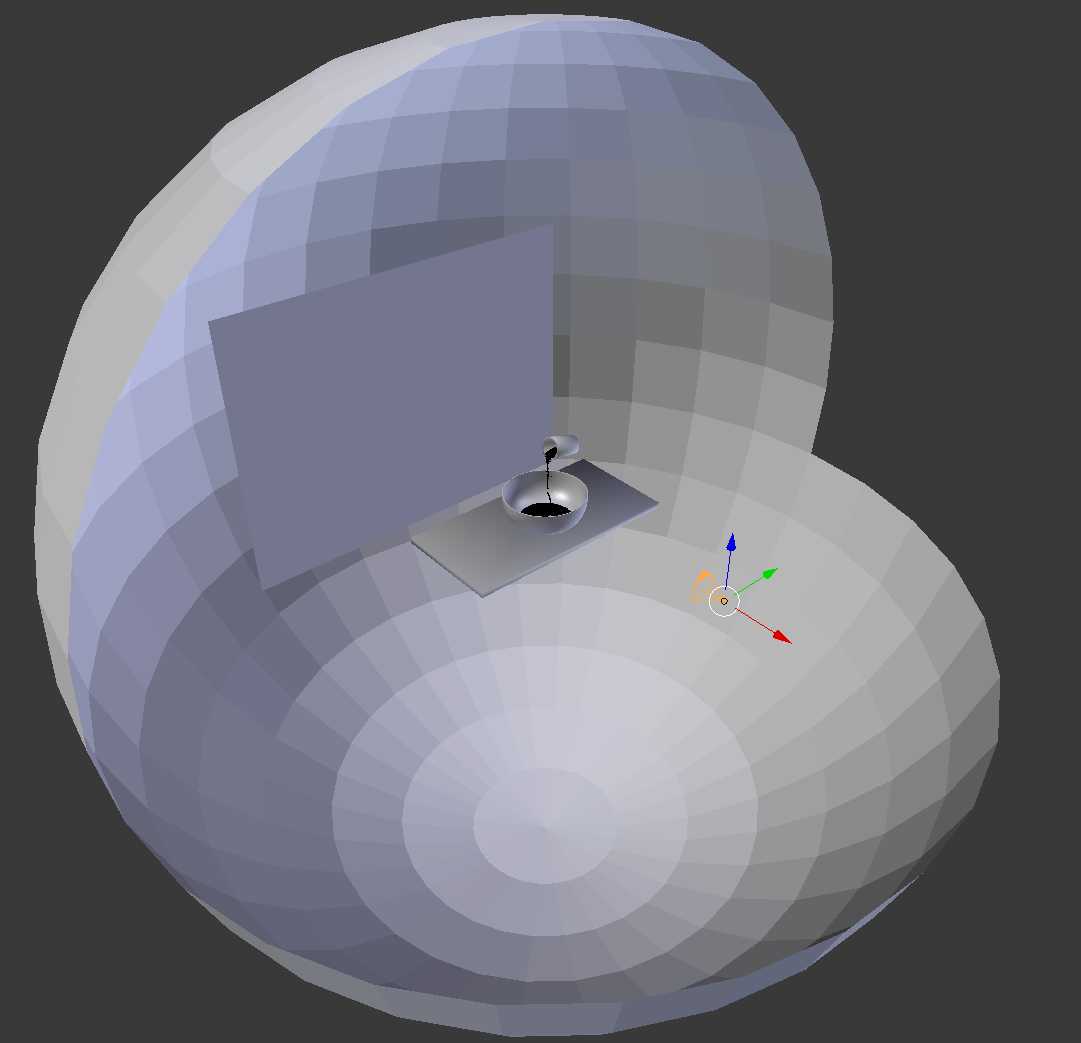}}}
    \end{picture}
    \caption{The setup used to simulate and render liquid sequences. The objects are shown here textureless for clarity. The sphere surrounding all the objects has been cut away to allow viewing of the objects inside. The orange shape represents the camera's viewpoint, and the flat plane across the table from it is the plane on which the video sequence is rendered. Note that this plane is sized to exactly fill the camera's view frustum. The background sphere is not directly visible by the camera and is used primarily to compute realistic reflections.}
    \label{fig:blender_scene}
    \vspace{-0.5cm}
\end{figure}

We generate data using the 3D-modeling application Blender \cite{blender2016} and the library El'Beem for liquid simulation, which is based on the lattice-Boltzmann method for efficient, physically accurate liquid simulations \cite{korner2006}. 
We separate the data generation process into two steps: simulation and rendering. 
During simulation, the liquid simulator calculates the trajectory of the surface mesh of the liquid as the cup pours the liquid into the bowl. 
We vary 4 variables during simulation: the type of cup (cup, bottle, mug), the type of bowl (bowl, dog dish, fruit bowl), the initial amount of liquid (30\% full, 60\% full, 90\% full), and the pouring trajectory (slow, fast, partial), for a total of 81 simulations. Each simulation lasts exactly 15 seconds for a total of 450 frames (30 frames per second).

\begin{figure}[t]
    \centering
    \setlength{\fboxsep}{0pt}
    \setlength{\fboxrule}{1pt}
    \setlength{\unitlength}{1.0cm}
    \begin{picture}(12.0,7.0)
        \put(0.0,0.0){\fbox{\includegraphics[width=3.0cm]{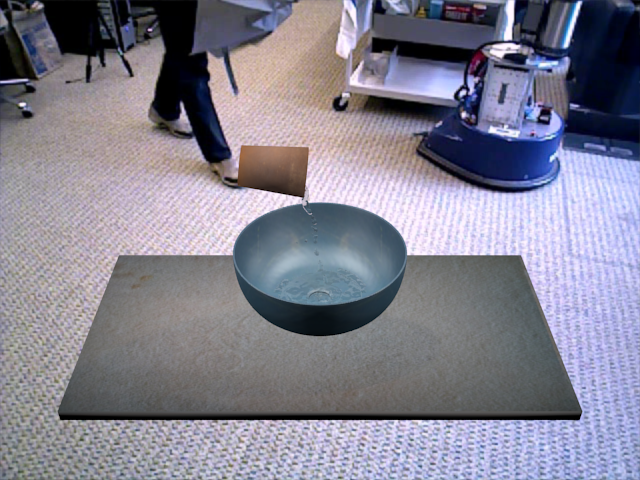}}}
        \put(3.0,0.0){\fbox{\includegraphics[width=3.0cm]{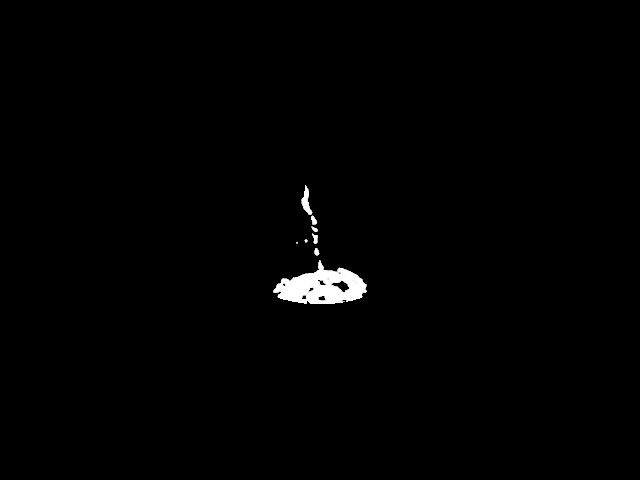}}}
        \put(6.0,0.0){\fbox{\includegraphics[width=3.0cm]{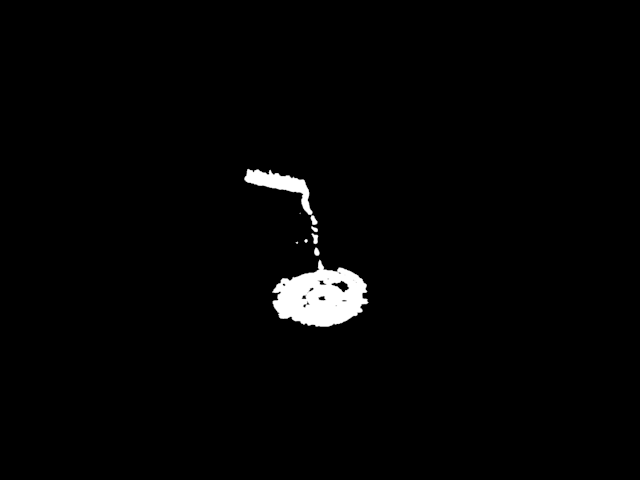}}}
        \put(9.0,0.0){\fbox{\includegraphics[width=3.0cm]{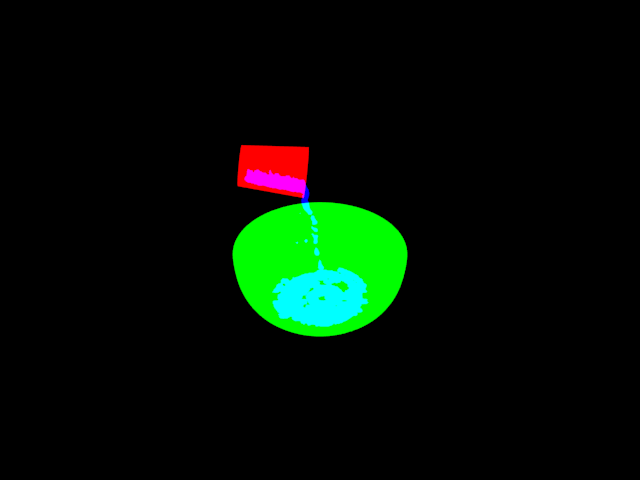}}}
        
        \put(0.0,2.25){\fbox{\includegraphics[width=3.0cm]{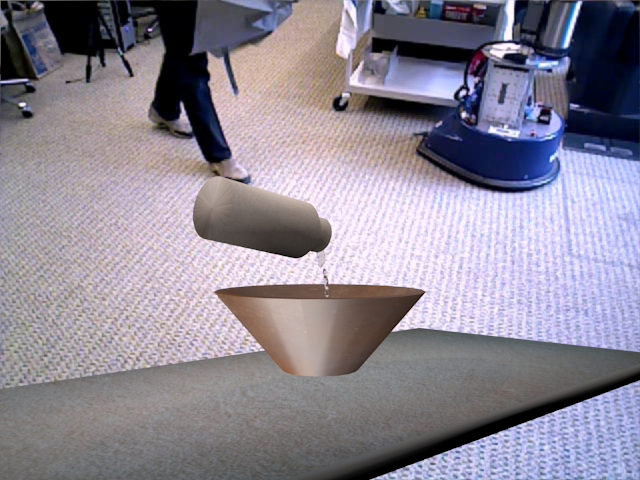}}}
        \put(3.0,2.25){\fbox{\includegraphics[width=3.0cm]{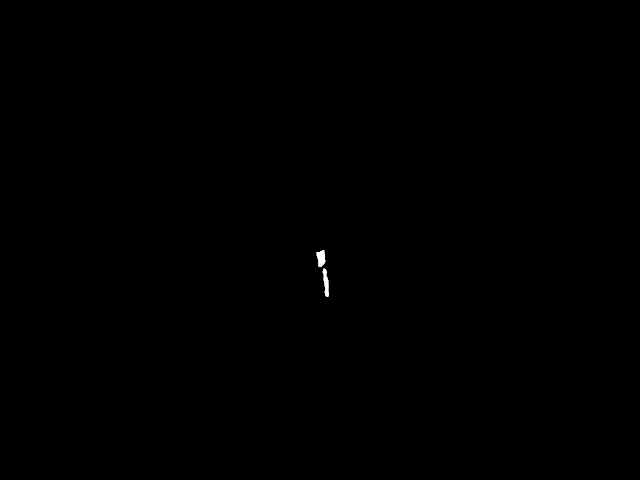}}}
        \put(6.0,2.25){\fbox{\includegraphics[width=3.0cm]{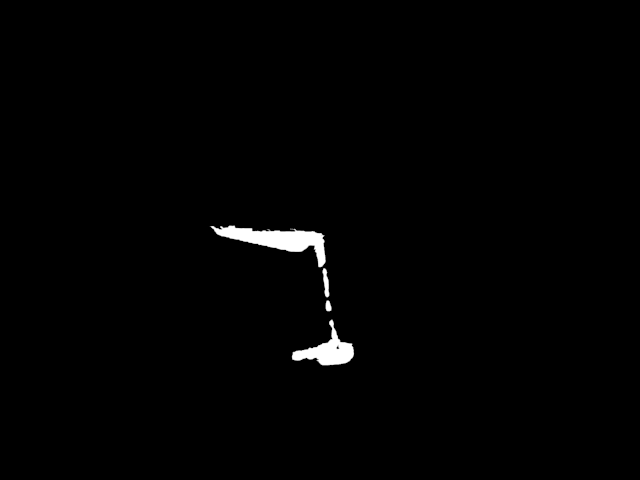}}}
        \put(9.0,2.25){\fbox{\includegraphics[width=3.0cm]{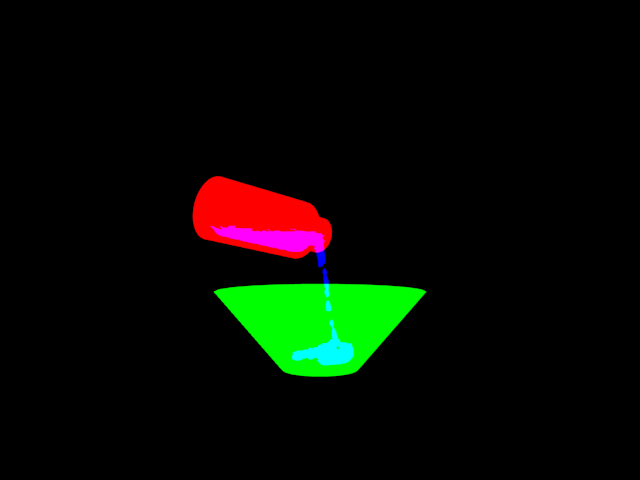}}}

        \put(0.0,4.5){\fbox{\includegraphics[width=3.0cm]{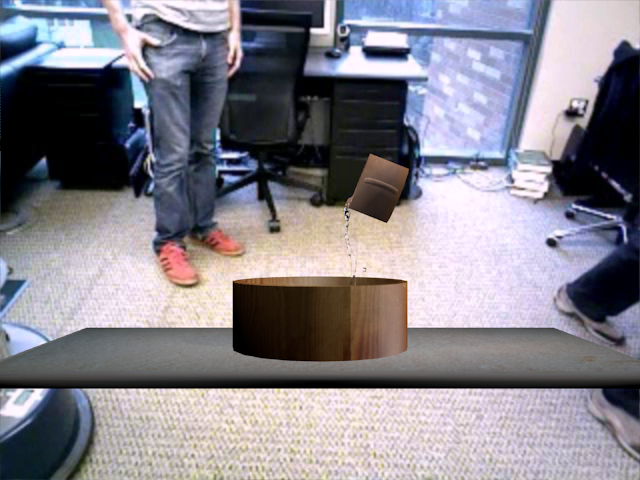}}}
        \put(3.0,4.5){\fbox{\includegraphics[width=3.0cm]{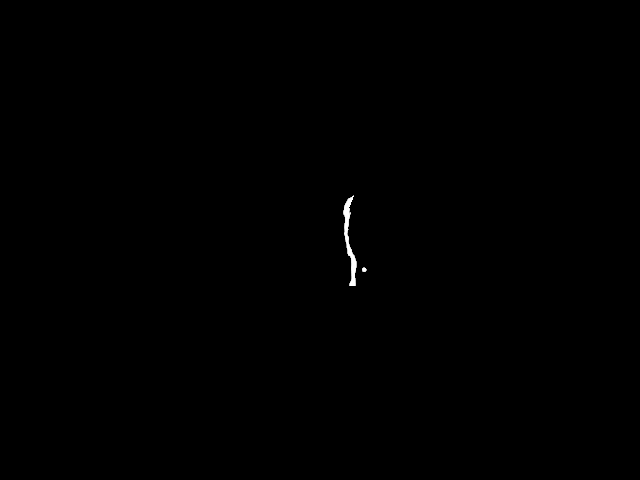}}}
        \put(6.0,4.5){\fbox{\includegraphics[width=3.0cm]{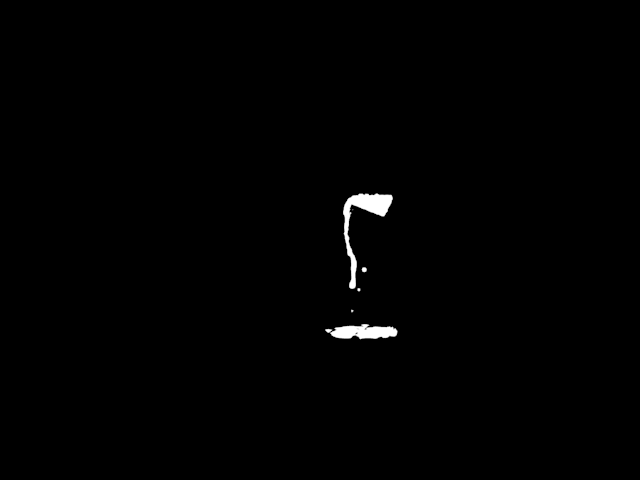}}}
        \put(9.0,4.5){\fbox{\includegraphics[width=3.0cm]{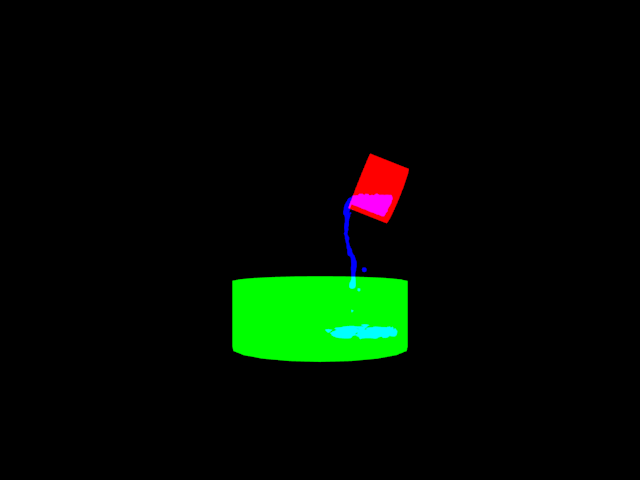}}}
        
        \put(1.0,6.85){{\bf RGB}}
        \put(3.8,6.85){{\bf Detection}}
        \put(6.8,6.85){{\bf Tracking}}
        \put(9.8,6.85){{\bf Labels}}
    \end{picture}
    \caption{Examples of frames rendered by our data generation algorithm. The left column is the raw RGB images generated by the renderer; the center-left column shows the ground truth liquid location for detection; the center-right column shows the ground truth liquid location for tracking; the right column shows the ground truth labeling output by the simulator.}
    \label{fig:data_gen}
    \vspace{-0.5cm}
\end{figure}
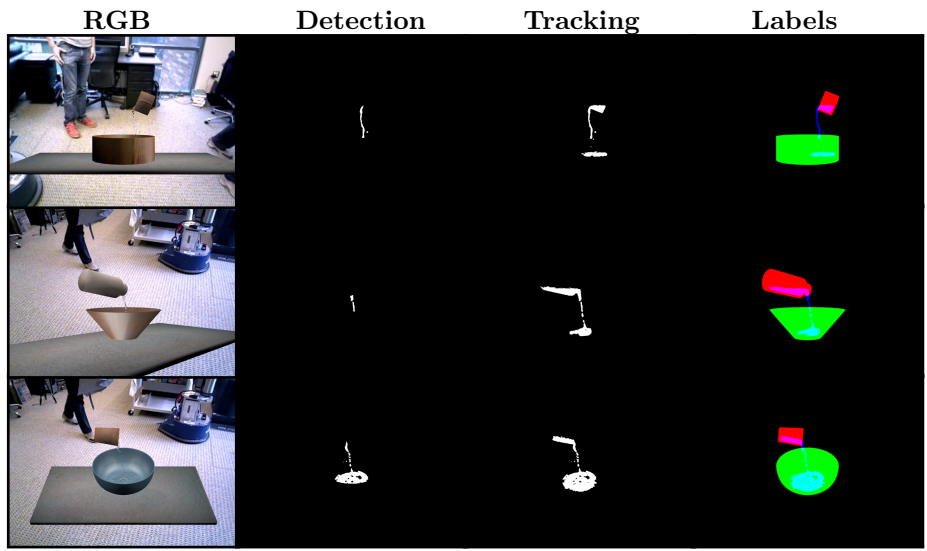

Next we render each simulation. 
We separate simulation from rendering because it allows us to vary other variables that don't affect the trajectory of the liquid mesh (e.g., camera viewpoint), which provides a significant speedup as liquid simulation is much more computationally intensive than rendering. 
In order to approximate realistic reflections, we mapped a 3D photo sphere image taken in our lab to the inside of a sphere, which we place in the scene surrounding all the objects. 
To prevent overfitting to a static background, we also add a plane in the image in front of the camera and behind the objects that plays a video of activity in our lab that approximately matches with that location in the background sphere. 
This setup is shown in Fig. \ref{fig:blender_scene}.
The liquid is always rendered as 100\% transparent, with only reflections, refractions, and specularities differentiating it from the background. 
For each simulation, we vary 6 variables: camera viewpoint (48 preset viewpoints), background video (8 videos), cup and bowl textures (6 textures each), liquid reflectivity (normal, none), and liquid index-of-refraction (air-like, low-water, normal-water). 
The 48 camera viewpoints were generated by varying the camera elevation (level with the table and looking down at a 45 degree angle), camera distance (8m, 10m, and 12m), and the camera azimuth (the 8 points of the compass, with north, southwest, and south shown in the top, middle, and bottoms rows of Fig. \ref{fig:data_gen}) respectively. 
We also generate negative examples without liquid. 
In total, this yields 165,888 possible renders for each simulation. 
It is infeasible to render them all, so we randomly sample variable values to render.

The labels are generated for each object (liquid, cup, bowl) as follows.
First, all other objects in the scene are set to render as invisible.
Next, the material for the object is set to render as a specific, solid color, ignoring lighting.
The sequence is then rendered, yielding a class label for the object for each pixel.
An example of labeled data (right column) and its corresponding rendered image (left column) is shown in Fig. \ref{fig:data_gen}. 
The cup, bowl, and liquid are rendered as red, green and blue respectively.
Note that this method allows each pixel to have multiple labels, e.g., some of the pixels in the cup are labeled as both cup and liquid (magenta in the right column of Fig. \ref{fig:data_gen}).
To determine which of the objects, if any, is visible at each pixel, we render the sequence once more with all objects set to render as their respective colors, and we use the alpha channel in the ground truth images to encode the visible class label. 

To evaluate our learning architectures, we generated 10,122 pouring sequences by randomly selecting render variables as described above as well as generating negative sequences (i.e., sequences without any water), for a total of 4,554,900 training images. Both the model files generated by Blender and the rendered images for the entire dataset are available for download at the following link: \url{http://rse-lab.cs.washington.edu/lpd/}.

\vspace{-0.3cm}
\subsection{Network Architecture}
\vspace{-0.3cm}

We test three network layouts for the tasks of detecting and tracking liquids: CNN, MF-CNN, and LSTM-CNN. All of our networks are fully-convolutional \cite{long2015}, that is, there are no fully-connected layers. In place of fully-connected layers used in more standard CNNs, we use $1 \times 1$ convolutional layers, which have a similar effect but prevent the explosion of parameters that normally occurs. We use the Caffe deep learning framework \cite{jia2014} to implement our networks\footnote{The network structure files (prototxt) can be found on our project page at \url{http://rse-lab.cs.washington.edu/projects/liquids/}}.

\begin{figure}[t]
\vspace{-0.5cm}
\begin{subfigure}{12.0cm}
    \begin{tikzpicture}[->,>=stealth,auto,node distance=1.6cm,thick,
      input node/.style={rectangle,draw,anchor=west,align=center,inner sep=0,outer sep=0},
      conv node/.style={rectangle,fill=red!60,draw,font=\sffamily\scriptsize\bfseries,align=center,anchor=west,minimum height=1.5cm},
      blob node/.style={ellipse,fill=gray!20,draw,font=\sffamily\scriptsize\bfseries,align=center,inner sep=0},
      elps node/.style={fill=none,draw=none,font=\sffamily\LARGE\bfseries}]

      \node[input node] (In3) at (0.0,4.0) {\includegraphics[width=2cm]{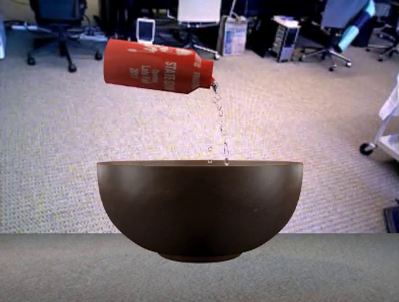}};

      \node[conv node] (Conv1) at (2.2,3.975) {\rotatebox{270}{\parbox[c]{1.45cm}{\centering Convolution\\{\normalfont\tiny\it 32 $5{\times}5$ kernels}}}};
      \node[conv node] (Conv2) at (3.1,3.975) {\rotatebox{270}{\parbox[c]{1.45cm}{\centering Convolution\\{\normalfont\tiny\it 32 $5{\times}5$ kernels}}}};
      \node[conv node] (Conv3) at (4.0,3.975) {\rotatebox{270}{\parbox[c]{1.45cm}{\centering Convolution\\{\normalfont\tiny\it 32 $5{\times}5$ kernels}}}};
      \node[conv node] (Conv4) at (4.9,3.975) {\rotatebox{270}{\parbox[c]{1.45cm}{\centering Convolution\\{\normalfont\tiny\it 32 $5{\times}5$ kernels}}}};
      \node[conv node] (Conv5) at (5.8,3.975) {\rotatebox{270}{\parbox[c]{1.7cm}{\centering Convolution\\{\normalfont\tiny\it 32 $17{\times}17$ kernels}}}};
      
      \node[conv node] (fc_conv1) at (6.9,3.975) [fill=blue!60]{\rotatebox{270}{\parbox[c]{2.0cm}{\centering $\mathbf{1{\times}1}$ Convolution\\{\normalfont\tiny\it 64 kernels}}}};
      \node[conv node] (fc_conv2) at (7.9,3.975) [fill=blue!60]{\rotatebox{270}{\parbox[c]{2.0cm}{\centering $\mathbf{1{\times}1}$ Convolution\\{\normalfont\tiny\it 64 kernels}}}};
      \node[conv node] (deconv) at (9.0,3.975) [fill=orange!60]{\rotatebox{270}{\parbox[c]{2.0cm}{\centering Deconvolution\\{\normalfont\tiny\it 64 $16{\times}16$ kernels}}}};
      \node[input node] (Out1) at (10.2,3.975) {\includegraphics[width=2cm]{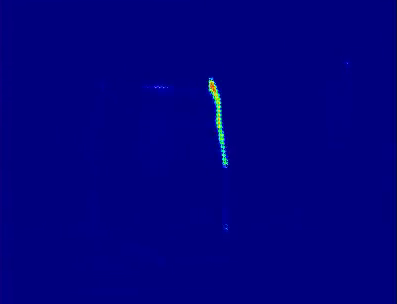}};
      
      \draw (In3) -- (Conv1);
      \draw (Conv1) -- (Conv2);
      \draw (Conv2) -- (Conv3);
      \draw (Conv3) -- (Conv4);
      \draw (Conv4) -- (Conv5);

      \draw (Conv5) -- (fc_conv1);
      
      \draw (fc_conv1) -- (fc_conv2);
      \draw (fc_conv2) -- (deconv);
      \draw (deconv) -- (Out1);
    \end{tikzpicture}
    \caption{Single-frame CNN}
    \label{fig:network-cnn}
\end{subfigure}

\begin{subfigure}{12.0cm}
    \begin{tikzpicture}[->,>=stealth,auto,node distance=1.6cm,thick,
      input node/.style={rectangle,draw,anchor=west,align=center,inner sep=0,outer sep=0},
      conv node/.style={rectangle,fill=red!60,draw,font=\sffamily\scriptsize\bfseries,align=center,anchor=west,minimum height=1.5cm},
      blob node/.style={ellipse,fill=gray!20,draw,font=\sffamily\scriptsize,align=center,inner sep=0},
      elps node/.style={fill=none,draw=none,font=\sffamily\LARGE\bfseries}]

      \node[input node] (In3) at (0.0,4.0) {\includegraphics[width=2cm]{graph_input.png}};
      \node[input node] (Rec1) at (0.0,2.225) {\includegraphics[width=2cm]{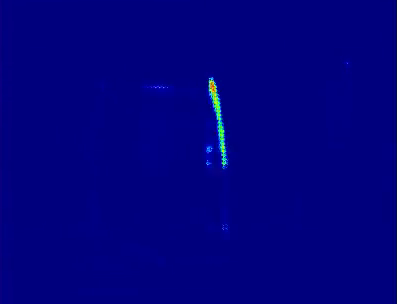}};

      \node[conv node] (Conv1) at (2.2,3.975) {\rotatebox{270}{\parbox[c]{1.45cm}{\centering Convolution\\{\normalfont\tiny\it 32 $5{\times}5$ kernels}}}};
      \node[conv node] (Conv2) at (3.1,3.975) {\rotatebox{270}{\parbox[c]{1.45cm}{\centering Convolution\\{\normalfont\tiny\it 32 $5{\times}5$ kernels}}}};
      \node[conv node] (Conv3) at (4.0,3.975) {\rotatebox{270}{\parbox[c]{1.45cm}{\centering Convolution\\{\normalfont\tiny\it 32 $5{\times}5$ kernels}}}};
      \node[conv node] (Conv4) at (4.9,3.975) {\rotatebox{270}{\parbox[c]{1.45cm}{\centering Convolution\\{\normalfont\tiny\it 32 $5{\times}5$ kernels}}}};
      \node[conv node] (Conv5) at (5.8,3.975) {\rotatebox{270}{\parbox[c]{1.7cm}{\centering Convolution\\{\normalfont\tiny\it 32 $17{\times}17$ kernels}}}};
      
      \node[conv node] (rec_conv1) at (2.2,2.2) {\rotatebox{270}{\parbox[c]{1.45cm}{\centering Convolution\\{\normalfont\tiny\it 20 $5{\times}5$ kernels}}}};
      \node[conv node] (rec_conv2) at (3.1,2.2) {\rotatebox{270}{\parbox[c]{1.45cm}{\centering Convolution\\{\normalfont\tiny\it 20 $5{\times}5$ kernels}}}};
      \node[conv node] (rec_conv3) at (4.0,2.2) {\rotatebox{270}{\parbox[c]{1.45cm}{\centering Convolution\\{\normalfont\tiny\it 20 $5{\times}5$ kernels}}}};
      
      \node[conv node] (lstm1) at (7.0,3.0) [fill=green!60]{\parbox[c]{0.65cm}{\centering LSTM\\{\normalfont\tiny\it 20~$1{\times}1$ kernels \\\vspace{-0.1cm}per~gate}}};
      \node[conv node] (fc_conv1) at (8.4,3.0) [fill=blue!60]{\rotatebox{270}{\parbox[c]{2.0cm}{\centering $\mathbf{1{\times}1}$ Convolution\\{\normalfont\tiny\it 64 kernels}}}};
      \node[conv node] (deconv) at (9.3,3.0) [fill=orange!60]{\rotatebox{270}{\parbox[c]{2.0cm}{\centering Deconvolution\\{\normalfont\tiny\it 64 $16{\times}16$ kernels}}}};
      \node[input node] (Out1) at (10.2,3.0) {\includegraphics[width=2cm]{graph_output.png}};

      \node[blob node] (rec_in2) at (6.0, 2.0) {Recurrent\\State};
      \node[blob node] (rec_in3) at (7.45, 1.7) {Cell\\State};
      
      \node[blob node] (rec_out2) at (8.7, 4.5) {Recurrent\\State};
      \node[blob node] (rec_out3) at (7.45, 4.3) {Cell\\State};
      
      \draw (In3) -- (Conv1);
      \draw (Conv1) -- (Conv2);
      \draw (Conv2) -- (Conv3);
      \draw (Conv3) -- (Conv4);
      \draw (Conv4) -- (Conv5);
      
      \draw (Rec1) -- (rec_conv1);
      \draw (rec_conv1) -- (rec_conv2);
      \draw (rec_conv2) -- (rec_conv3);

      \draw (Conv5.east) -- (lstm1.west);
      \draw (rec_conv3.east) -- (lstm1.west);
      
      \draw (lstm1) -- (fc_conv1);
      \draw (fc_conv1) -- (deconv);
      \draw (deconv) -- (Out1);
      
      \draw (rec_in2) -- (lstm1.west);

      \draw (rec_in3) -- (lstm1.south);
      \draw (lstm1.east) -- (rec_out2.215);
      \draw (lstm1.north) -- (rec_out3);
    \end{tikzpicture}
    \caption{LSTM-CNN}
    \label{fig:network-lstm}
\end{subfigure}
\caption{Layout of the single-frame and LSTM networks. Each of the red {\bf Convolution} layers is followed by a max pooling layer and a rectified linear layer. The max pooling layers have the same kernel size as the convolution layer they follow, and the first two convolution layers in each network have a stride of 2 (all others and all convolution layers have a stride of 1). Each of the blue {\bf $\mathbf{1{\times}1}$ Convolution} layers is followed by a rectified linear layer. Refer to figure 1 of \cite{greff2015} for more details on the LSTM layer.}
\label{fig:network}
\vspace{-0.5cm}
\end{figure}

\vspace{-0.5cm}
\subsubsection{CNN} 

The first layout is a standard convolutional neural network (CNN). It takes in an image and outputs probabilities for each class label at each pixel. It has a fixed number of convolutional layers, each followed by a rectified linear layer and a max pooling layer. In place of fully-connected layers, we use two $1 \times 1$ convolutional layers, each followed by a rectified linear layer. The last layer of the network is a deconvolutional layer that upsamples the output of the $1 \times 1$ convolutional layers to be the same size as the input image. This network is shown in Fig. \ref{fig:network-cnn}.

\vspace{-0.5cm}
\subsubsection{MF-CNN} 

The second layout is a multi-frame CNN. Instead of taking in a single frame, it takes as input multiple consecutive frames and predicts the probability of each class label for each pixel at the last frame. It is similar to the single-frame CNN network shown in Fig. \ref{fig:network-cnn} except each frame is convolved independently through the first 5 convolution layers, and then the output for each frame is concatenated together channel-wise. This is fed to the two $1 \times 1$ convolutional layers, each followed by a rectified linear layer, and finally a deconvolutional layer. We fix the number of input frames for this layout to 32 for this paper, i.e., approximately 1 second's worth of data (30 frames per second), which we empirically determined strikes the best balance between window size and memory utilization.

\vspace{-0.5cm}
\subsubsection{LSTM-CNN} 

The third layout is similar to the single frame CNN layout, with the first $1 \times 1$ convolutional layer replaced with a LSTM layer (see figure 1 of \cite{greff2015} for a detailed layout of the LSTM layer). We replace the fully-connected layers of a standard LSTM with $1 \times 1$ convolutional layers. The LSTM takes as recurrent input the cell state from the previous timestep, its output from the previous timestep, and the output of the network from the previous timestep processed through 3 convolutional layers (each followed by a rectified linear and max pooling layer). During training, when unrolling the LSTM-CNN, we initialize this last recurrent input with the ground truth at the first timestep, but during testing we use the standard recurrent network technique of initializing it with all zeros. Fig. \ref{fig:network-lstm} shows the layout of the LSTM-CNN. 

\vspace{-0.5cm}
\section{Evaluation}
\label{sec:evaluation}
\vspace{-0.3cm}

We evaluated our networks on 4 experiments: fixed-viewpoint detection, multi-viewpoint detection, fixed-viewpoint tracking, and combined detection \& tracking. We define the detection task as, given raw color images, determine where the {\it visible} liquid in the images is. We define the tracking task as, given segmented images (i.e., images that have already been run through a detector), determine where {\it all} liquid (visible and occluded) is in the image. Intuitively, detection corresponds to perceiving the liquid, while tracking corresponds to reasoning about where the liquid is given what is (and has been) visible.

Every network was trained  using the mini-batch gradient descent method Adam \cite{kingma2014} with a learning rate of 0.0001 and default momentum values. Each network was trained for 61,000 iterations, at which point performance tended to plateau. All single-frame networks were trained using a batch size of 32; all multi-frame networks with a window of 32 and batch size of 1; and all LSTM networks with a batch size of 5. For all experiments except the third (fixed-viewpoint tracking), the input images were scaled to $400 \times 300$ resolution. The error signal was computed using the softmax with loss layer built into Caffe \cite{jia2014}.

We empirically determined, however, that naively training a network in this setup results in it predicting no liquid present in any scene at all due to the significant positive-negative class imbalance (most of the pixels in each image are non-liquid pixels). To counteract this we employed two strategies. The first was to pre-train the network on $160 \times 160$ crops of the image around liquid pixels. Since our networks are fully-convolutional, they can have variable sized inputs and outputs, which means a network pre-trained in this manner can be immediately trained on full images without needing any modification. The second strategy was to weight the gradients from the error signal based on the class of the ground truth pixel: 1.0 for positive pixels and 0.1 for negative pixels. This decreases the effect of the non-liquid pixels and prevents the network from predicting no liquid in the scene.

We report the precision and recall of each network on a hold-out test set, evaluated on pixel-wise classifications. We also report the precision and recall for various amounts of ``slack,'' i.e., we count a pixel labeled as liquid correct if it is within $n$ pixels of a ground truth liquid pixel, where $n$ is the amount of slack. This better evaluates the network in cases where it's predictions are only a few pixels off, which is a relatively small error given the resolution of the images.

\vspace{-0.3cm}
\subsection{Experiment 1: Fixed-Viewpoint Detection}
\vspace{-0.3cm}

We evaluated all three network types on a fixed-viewpoint detection task. 
We define fixed-viewpoint in this context to mean data generated as described in Section \ref{sec:data_gen} for which the camera elevation is level with the table and the azimuth is either north (as shown in the top row of Fig. \ref{fig:data_gen}) or south (180 degrees opposite). 
The networks were given the full rendered RGB image as input (similar to the left column in Fig. \ref{fig:data_gen}) and the output was a classification at each pixel as liquid or not liquid. 
To counteract the class imbalance, we employed visible liquid image crop pre-training for each network (we initialized the image crop LSTM-CNN with the trained weights of the image crop single-frame CNN). 
We then trained the final network for each type on full images initializing it with the weights of the image crop network. 
During training, the LSTM-CNN was unrolled for 32 timesteps.

\vspace{-0.3cm}
\subsection{Experiment 2: Multi-Viewpoint Detection}
\vspace{-0.3cm}

For the second experiment, we expanded the data used to include all 48 viewpoints, presenting a non-trivial increase in difficulty for the networks. 
Our goal was to test the generalizability of the networks across a much wider variation in viewpoints. 
For this reason, we focused only on testing the best performing network, the LSTM-CNN (see Section \ref{sec:results1} for results from experiment 1).
Also to test generalizability, we only trained the network on a subset of the 48 viewpoints, and tested on the remaining.
We used all data generated using the 8m and 12m camera viewpoint distances for training and data generated using the 10m camera distance for testing.  
We also employed the gradient weighting scheme described above to counteract the class imbalance.
The LSTM-CNN was trained in the same manner as in experiment 1.

\vspace{-0.3cm}
\subsection{Experiment 3: Fixed-Viewpoint Tracking}
\vspace{-0.3cm}

For tracking only, the networks were given pre-segmented input images, with the goal being to track the liquid when it is not visible. An example of this input is shown in the first row of the right column from Fig. \ref{fig:data_gen}, with the exception that the occluded liquid (magenta and cyan) were not shown. Because these input images are more structured, we lowered the resolution to $130 \times 100$. The output was the pixel-wise classification of liquid or not liquid, including pixels where the liquid was occluded by other objects in the scene. During training, the LSTM-CNN was unrolled for 160 timesteps. We reduced the number of initial convolution layers on the input from 5 to 3 for each of the three networks. Due to the structured nature of the input, each network was trained directly on full images with Gaussian-random weight initialization. We used the data from the same viewpoints (level with the table and azimuth at north or south) as in experiment 1.

\vspace{-0.3cm}
\subsection{Experiment 4: Combined Detection \& Tracking}
\vspace{-0.3cm}

For the last experiment, we combine detection and tracking into a single task, i.e., given raw color images, determine where {\it all} liquid in the scene is (visible and occluded). 
Our goal is to determine if it is possible to do both tasks with one network, and for this reason, we evaluate only the LSTM-CNN.
We initialized the network with the weights of the trained LSTM-CNN from experiment 1 and trained it on full images. 
As in experiment 2, we employed the gradient weighting scheme described above to counteract the class imbalance. 
We used the data from the same viewpoints as in experiment 1 and 3.

\vspace{-0.6cm}
\section{Results}
\label{sec:results}
\vspace{-0.3cm}
\subsection{Fixed-Viewpoint Detection}
\label{sec:results1}
\vspace{-0.2cm}

\begin{figure}[t]
    \centering
    \setlength{\fboxsep}{0pt}
    \setlength{\fboxrule}{1pt}
    \setlength{\unitlength}{1.0cm}
    \begin{picture}(10.0,7.7)
        \put(0.0,0.0){\fbox{\includegraphics[width=2.0cm]{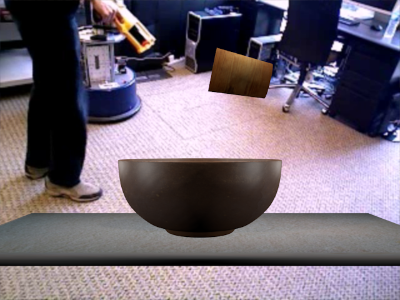}}}
        \put(2.0,0.0){\fbox{\includegraphics[width=2.0cm]{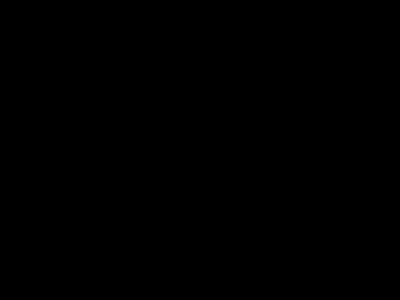}}}
        \put(4.0,0.0){\fbox{\includegraphics[width=2.0cm]{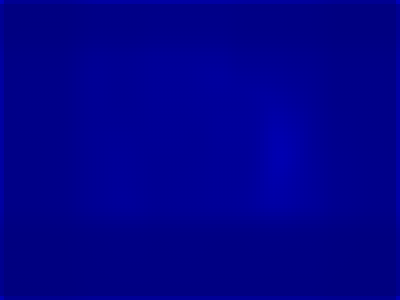}}}
        \put(6.0,0.0){\fbox{\includegraphics[width=2.0cm]{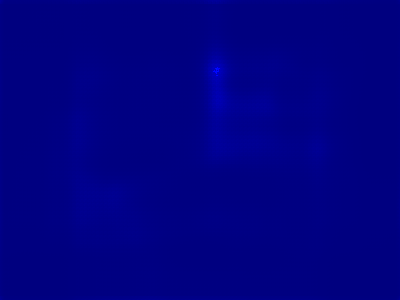}}}
        \put(8.0,0.0){\fbox{\includegraphics[width=2.0cm]{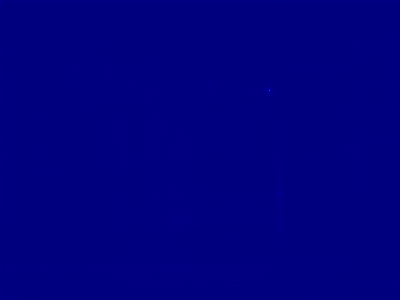}}}
        
        \put(0.0,1.55){\fbox{\includegraphics[width=2.0cm]{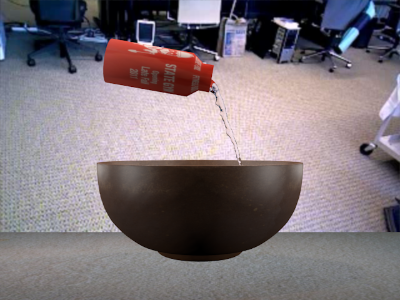}}}
        \put(2.0,1.55){\fbox{\includegraphics[width=2.0cm]{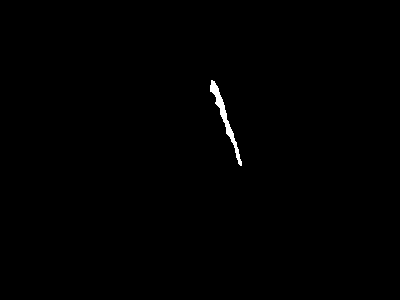}}}
        \put(4.0,1.55){\fbox{\includegraphics[width=2.0cm]{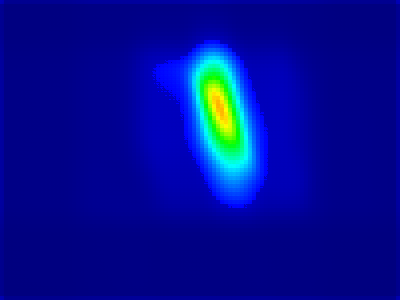}}}
        \put(6.0,1.55){\fbox{\includegraphics[width=2.0cm]{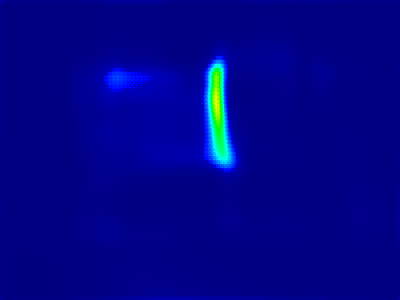}}}
        \put(8.0,1.55){\fbox{\includegraphics[width=2.0cm]{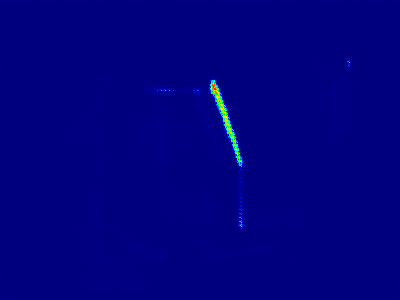}}}

        \put(0.0,3.1){\fbox{\includegraphics[width=2.0cm]{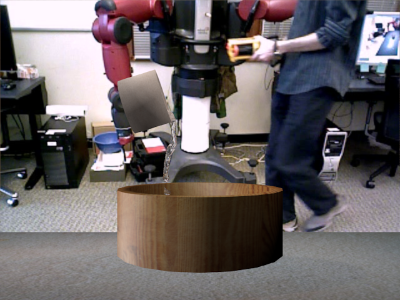}}}
        \put(2.0,3.1){\fbox{\includegraphics[width=2.0cm]{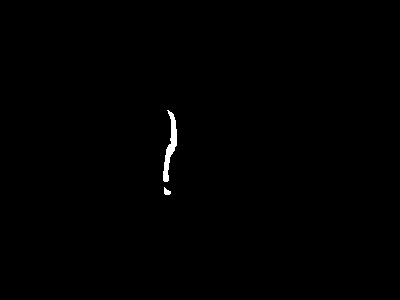}}}
        \put(4.0,3.1){\fbox{\includegraphics[width=2.0cm]{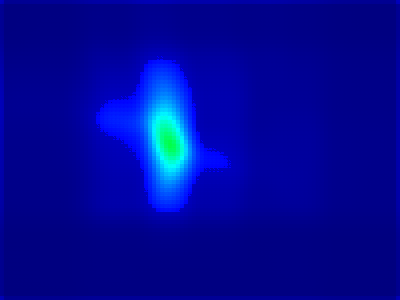}}}
        \put(6.0,3.1){\fbox{\includegraphics[width=2.0cm]{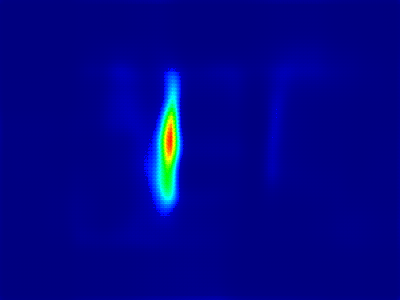}}}
        \put(8.0,3.1){\fbox{\includegraphics[width=2.0cm]{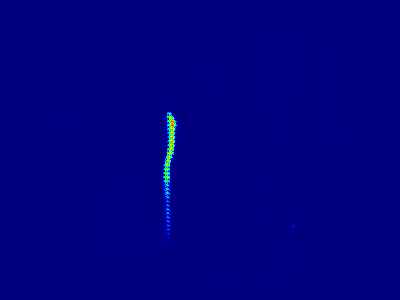}}}
        
        \put(0.0,4.65){\fbox{\includegraphics[width=2.0cm]{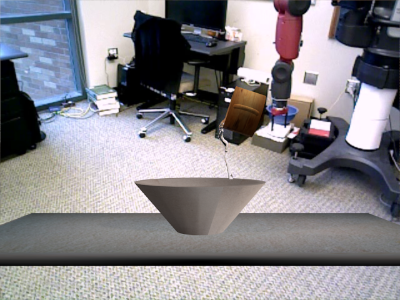}}}
        \put(2.0,4.650){\fbox{\includegraphics[width=2.0cm]{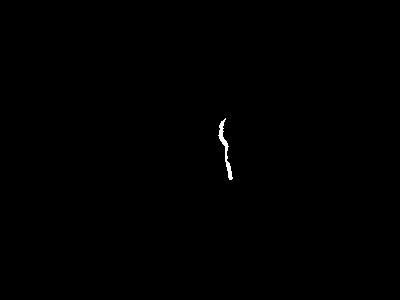}}}
        \put(4.0,4.65){\fbox{\includegraphics[width=2.0cm]{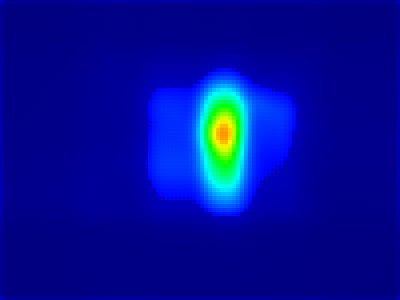}}}
        \put(6.0,4.65){\fbox{\includegraphics[width=2.0cm]{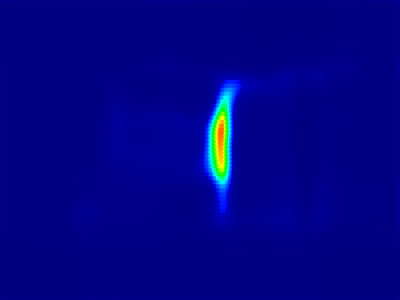}}}
        \put(8.0,4.65){\fbox{\includegraphics[width=2.0cm]{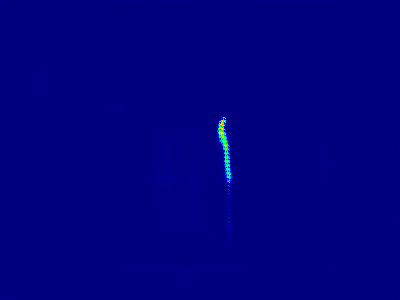}}}
        
        \put(0.0,6.2){\fbox{\includegraphics[width=2.0cm]{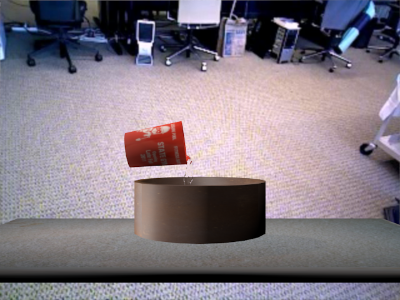}}}
        \put(2.0,6.2){\fbox{\includegraphics[width=2.0cm]{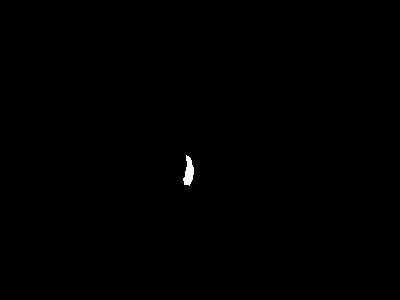}}}
        \put(4.0,6.2){\fbox{\includegraphics[width=2.0cm]{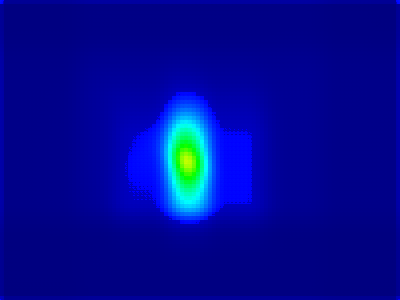}}}
        \put(6.0,6.2){\fbox{\includegraphics[width=2.0cm]{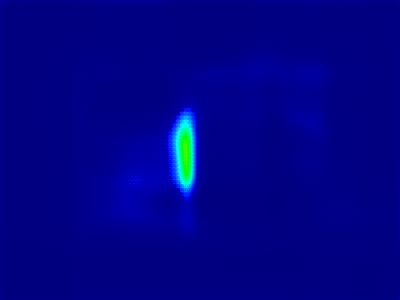}}}
        \put(8.0,6.2){\fbox{\includegraphics[width=2.0cm]{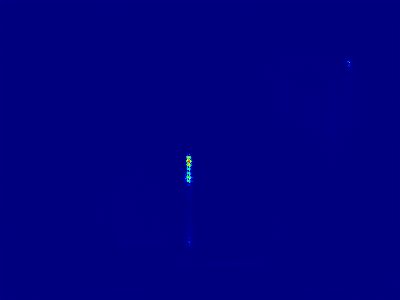}}}
        
        \put(0.5,7.9){{\bf Input}}
        \put(2.4,7.9){{\bf Labels}}
        \put(4.6,7.9){{\bf CNN}}
        \put(6.25,7.9){{\bf MF-CNN}}
        \put(8.05,7.9){{\bf LSTM-CNN}}
    \end{picture}
    \vspace{-0.2cm}
    \caption{Qualitative fixed-viewpoint liquid detection results. The Input column is the input to the networks, the Labels column is the ground truth labeling of each pixel as liquid or not liquid, and the CNN, MF-CNN, and LSTM-CNN columns show a heatmap of the prediction of each network for each of the input frames. 5 sequences were randomly selected from our training set, and the frame with the most liquid pixels was picked for display here, with the exception of the last row, which shows how the networks perform when there is no liquid present.}
    \label{fig:results}
    \vspace{-0.5cm}
\end{figure}
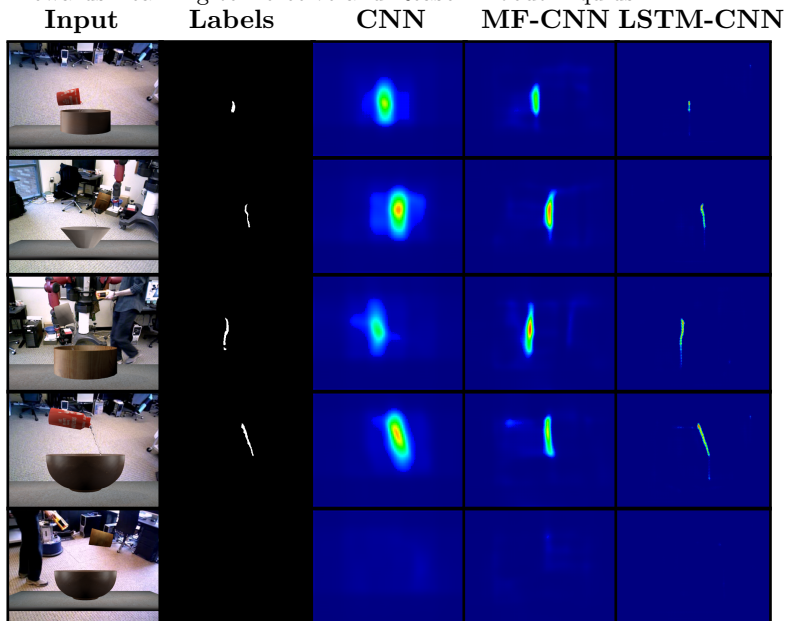

\begin{figure}
    \centering
    \setlength{\unitlength}{1.0cm}
    \begin{subfigure}{3.0cm}
        \begin{picture}(3.0,2.5)
            \put(0.0,0.0){\includegraphics[width=3.0cm]{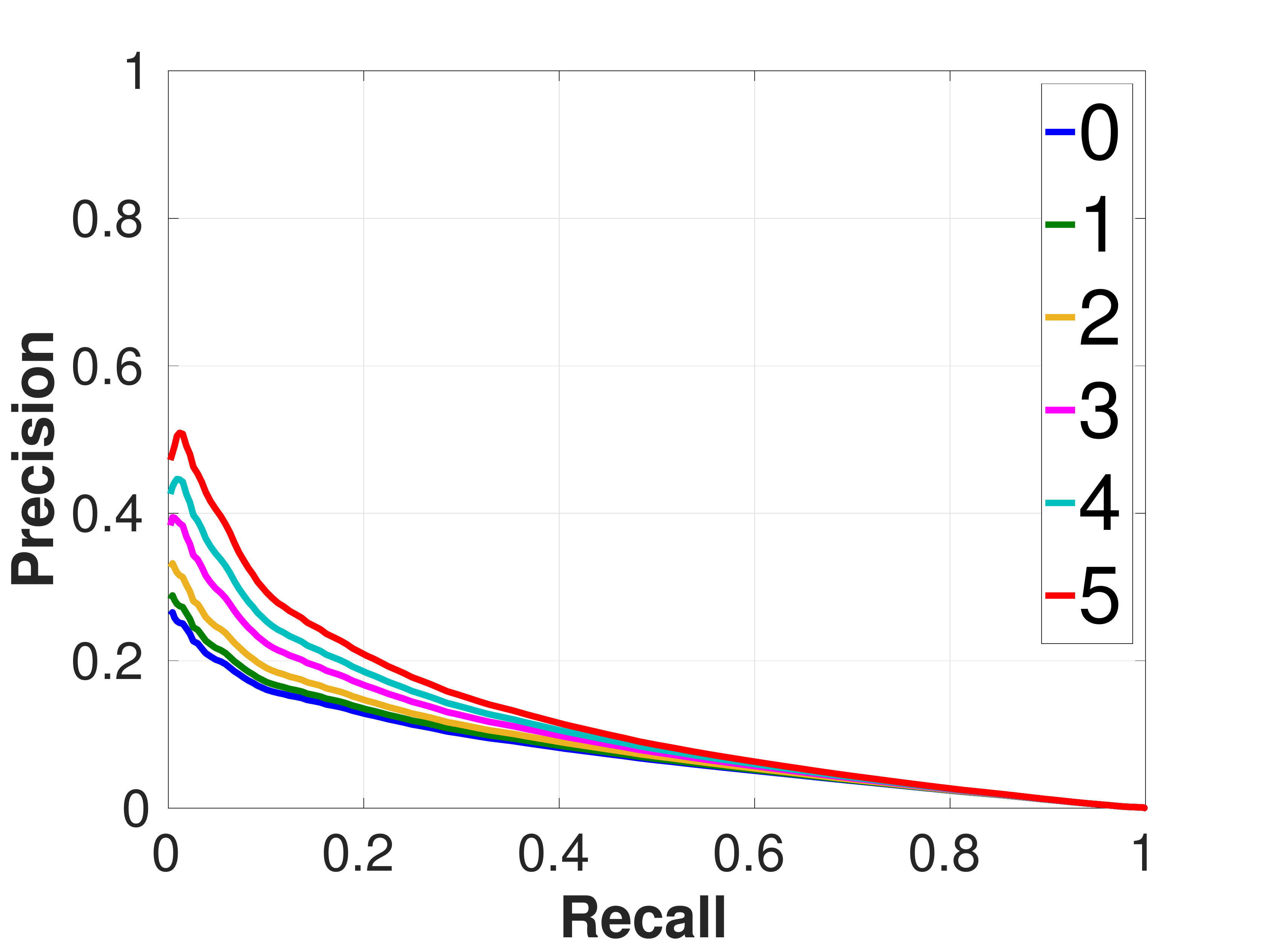}}
        \end{picture}
        \caption{CNN}
        \label{fig:results_detection_cnn}
    \end{subfigure}%
    \begin{subfigure}{3.0cm}
        \begin{picture}(3.0,2.5)
            \put(0.0,0.0){\includegraphics[width=3.0cm]{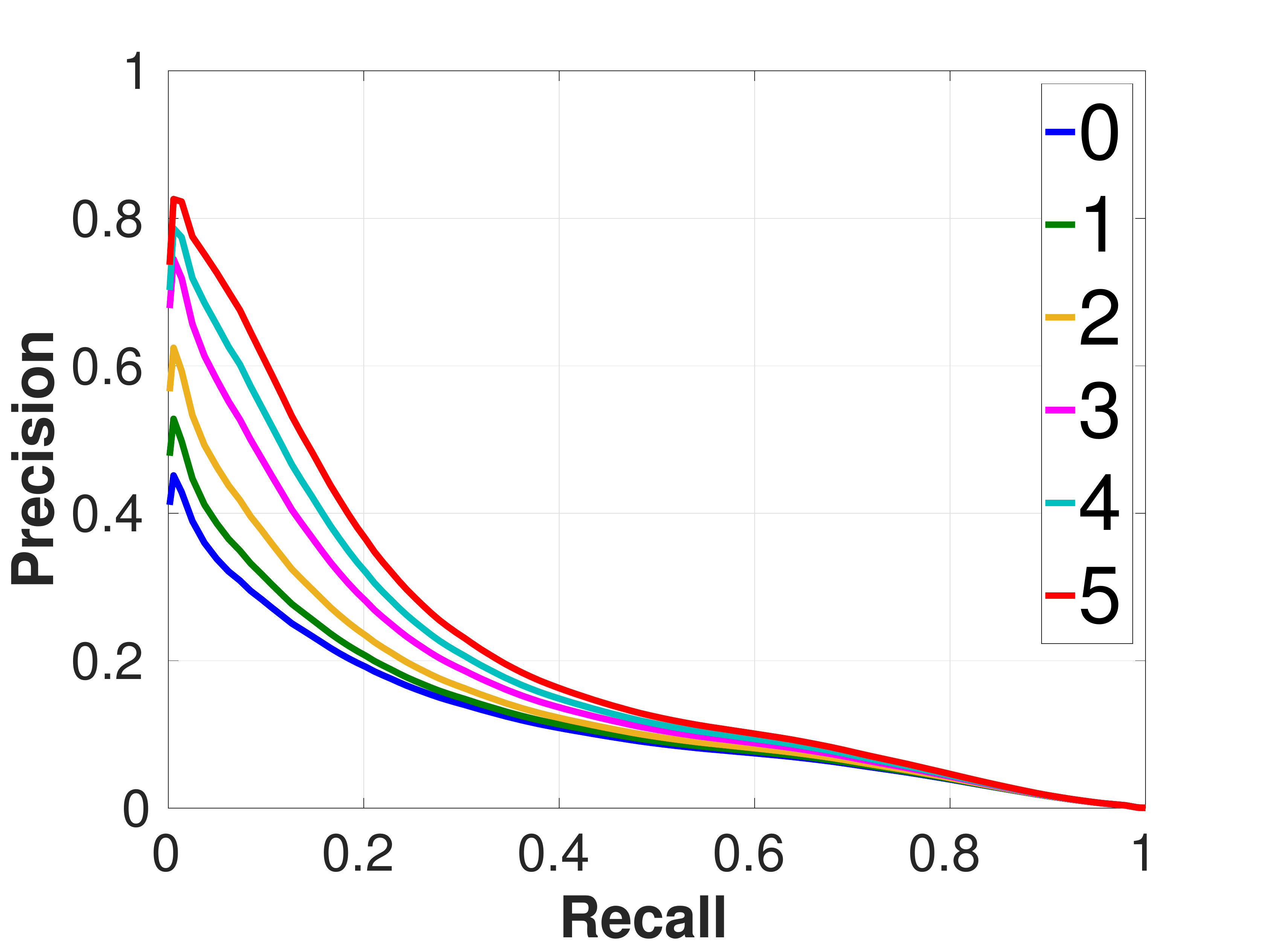}}
        \end{picture}
        \caption{MF-CNN}
        \label{fig:results_detection_wcnn}
    \end{subfigure}%
    \begin{subfigure}{3.0cm}
        \begin{picture}(3.0,2.5)
            \put(0.0,0.0){\includegraphics[width=3.0cm]{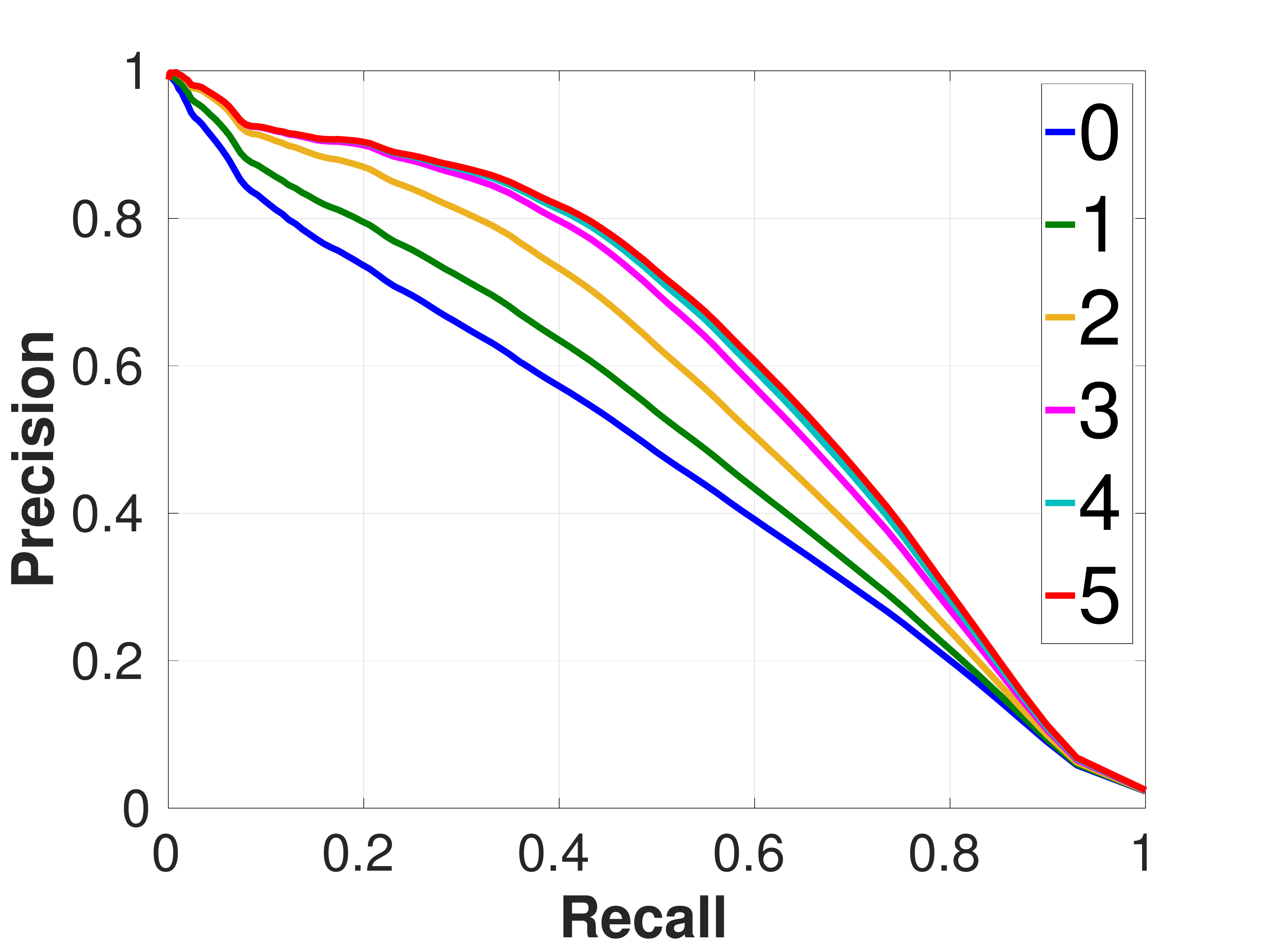}}
        \end{picture}
        \caption{LSTM-CNN}
        \label{fig:results_detection_lstm}
    \end{subfigure}%
    \begin{subfigure}{3.0cm}
        \begin{picture}(3.0,2.5)
            \put(0.0,0.0){\includegraphics[width=3.0cm]{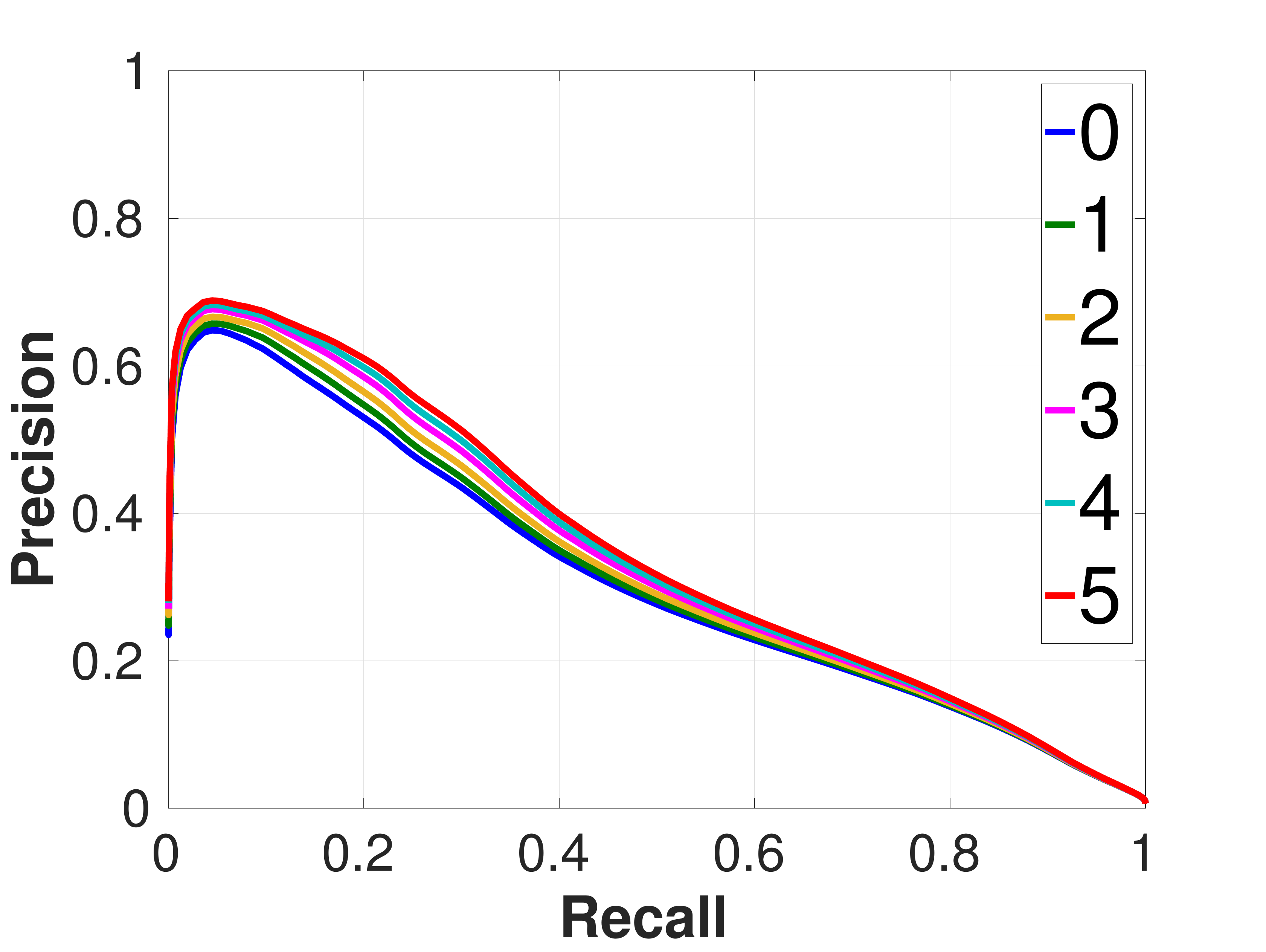}}
        \end{picture}
        \caption{MV LSTM-CNN}
        \label{fig:results_multiview_detection}
    \end{subfigure}
    \vspace{-0.3cm}
    \caption{Quantitative fixed- and multi-viewpoint liquid detection results. The graphs indicate the precision and recall for each of the three networks on fixed-viewpoint detection and the LSTM-CNN on multi-viewpoint detection. The colored lines indicate the variation in the number of slack pixels we allowed for prediction, i.e., how many pixels a positive classification could be away from a positive ground truth labeling and still be counted as correct.}
    \label{fig:results_detection}
    \vspace{-0.5cm}
\end{figure}

Fig. \ref{fig:results} shows qualitative results for the three networks on the liquid detection task\footnote{Video of the full sequences at \url{\youtubeurl}}. The frames in this figure were randomly selected from the training set, and it is clear from the figure that all three networks detect the liquid at least to some degree. \Cref{fig:results_detection_cnn,fig:results_detection_wcnn,fig:results_detection_lstm} show a quantitative comparison between the three networks. As expected, the multi-frame CNN outperforms the single-frame. Surprisingly, the LSTM-CNN performs much better than both by a significant margin. These results strongly suggest that detecting transparent liquid must be done over a series of frames, rather than a single frame.

\vspace{-0.3cm}
\subsection{Multi-Viewpoint Detection}
\vspace{-0.3cm}

Fig. \ref{fig:results_multiview_detection} shows the results from multi-viewpoint detection for the LSTM-CNN. As expected, the 8-fold increase in number of viewpoints leads to lower performance as compared to Fig. \ref{fig:results_detection_lstm}, but overall it is clearly still able to detect the liquid reasonably well. Interestingly, there is less spread between the various levels of slack than in Fig. \ref{fig:results_detection_lstm}, meaning the network benefits less from increased slack, suggesting that it is less precise than the fixed-view LSTM-CNN, which makes sense given the much larger variation in viewpoints.

\vspace{-0.5cm}
\subsection{Fixed-Viewpoint Tracking}
\label{sec:results3}
\vspace{-0.3cm}

\begin{figure}
    \vspace{-0.5cm}
    \centering
    \setlength{\unitlength}{1.0cm}
    \begin{subfigure}{3.0cm}
        \begin{picture}(3.0,2.7)
            \put(0.0,0.0){\includegraphics[width=3.0cm]{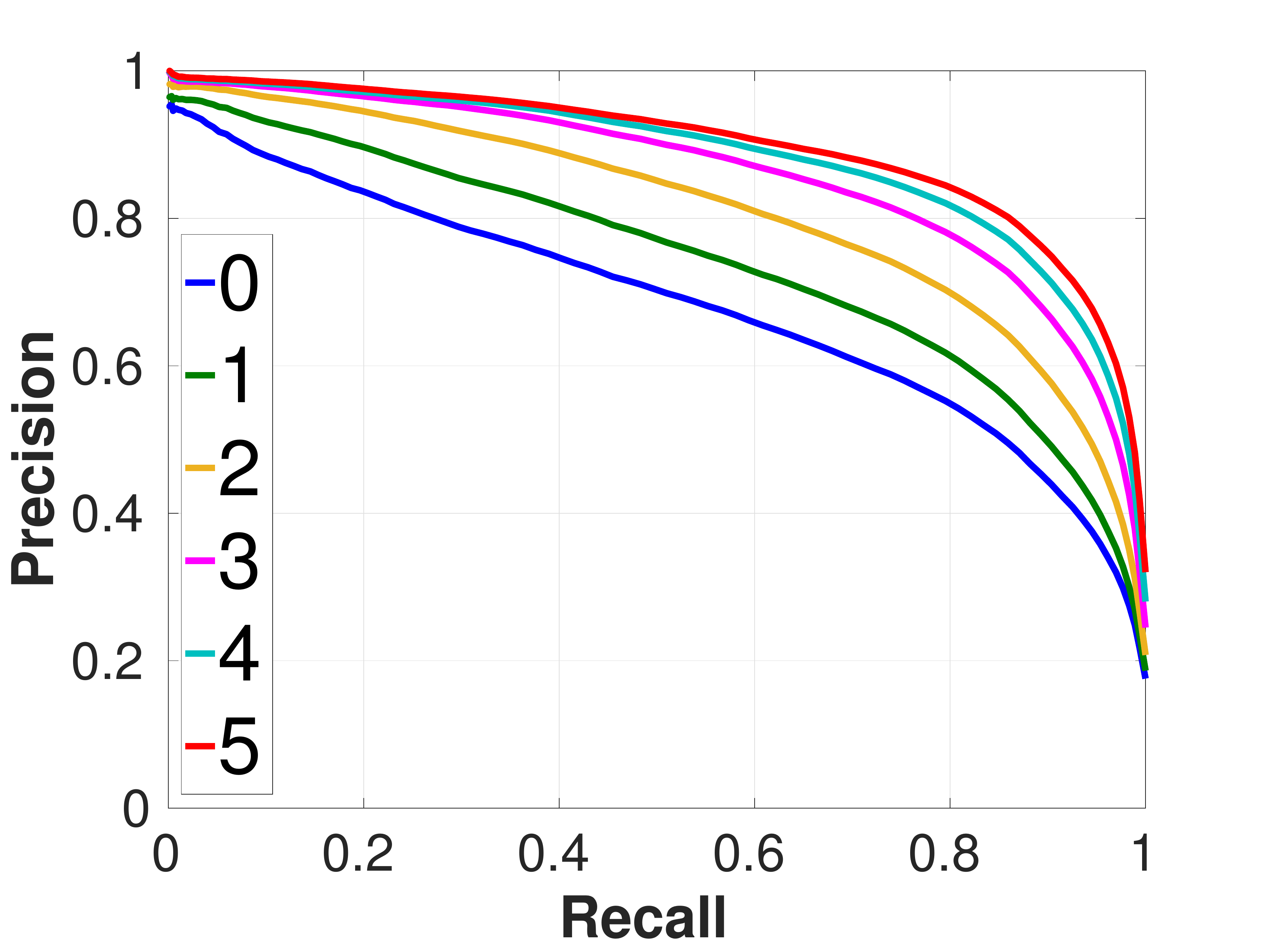}}
        \end{picture}
        \caption{CNN}
        \label{fig:results_tracking_cnn}
    \end{subfigure}%
    \begin{subfigure}{3.0cm}
        \begin{picture}(3.0,2.7)
            \put(0.0,0.0){\includegraphics[width=3.0cm]{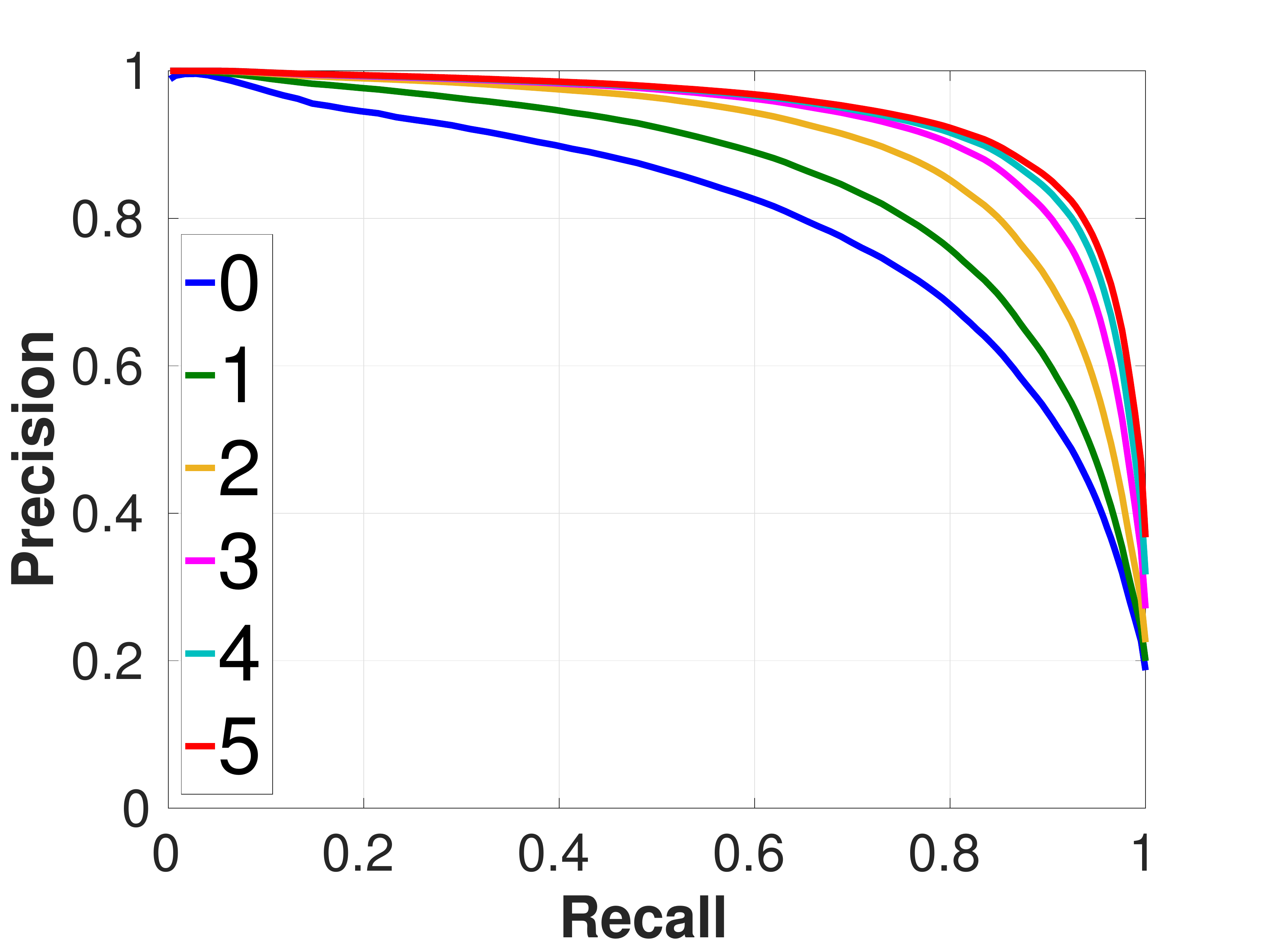}}
        \end{picture}
        \caption{MF-CNN}
        \label{fig:results_tracking_wcnn}
    \end{subfigure}%
    \begin{subfigure}{3.0cm}
        \begin{picture}(3.0,2.7)
            \put(0.0,0.0){\includegraphics[width=3.0cm]{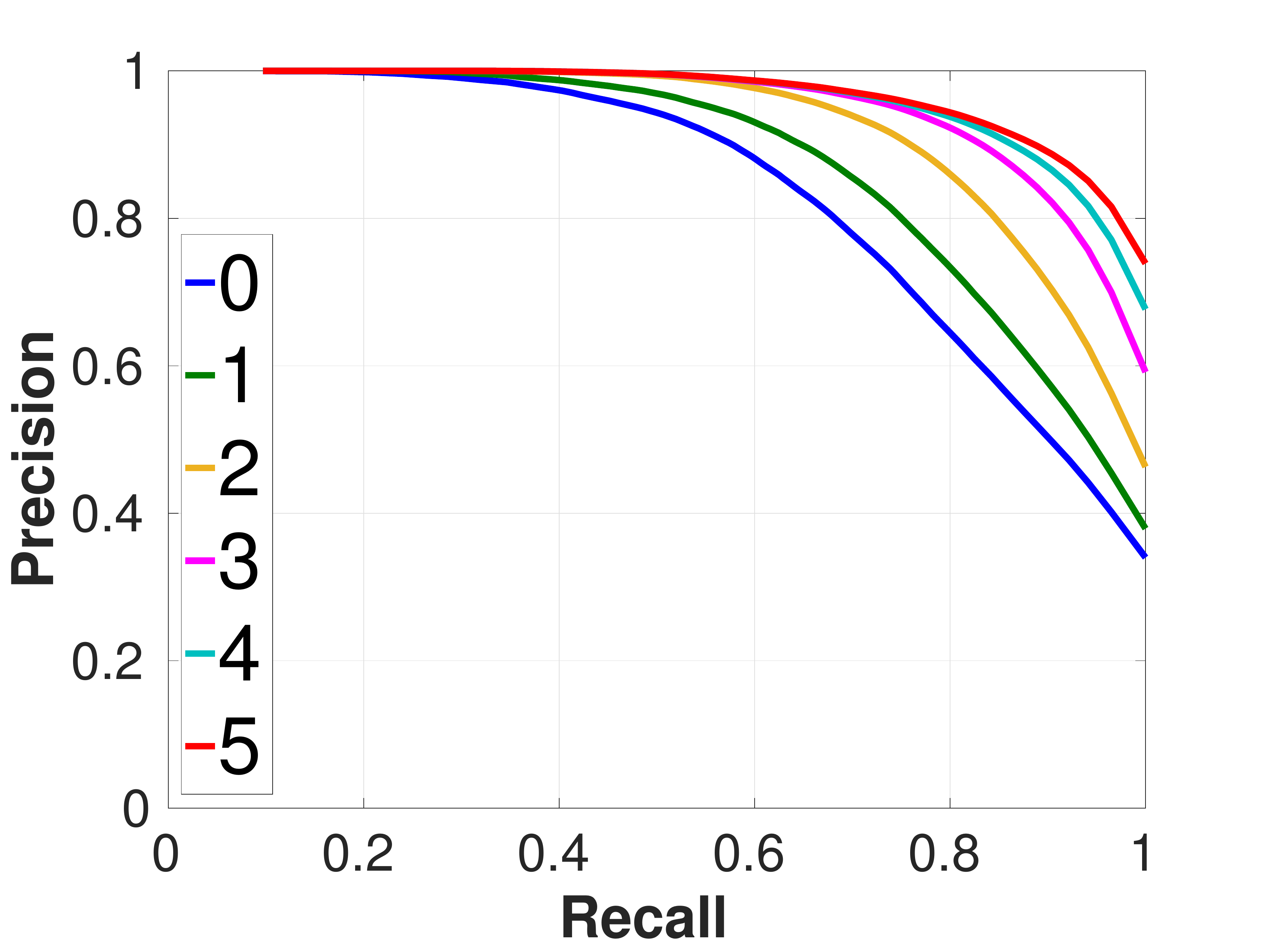}}
        \end{picture}
        \caption{LSTM-CNN}
        \label{fig:results_tracking_lstm}
    \end{subfigure}%
    \begin{subfigure}{3.0cm}
        \begin{picture}(3.0,2.7)
            \put(0.0,0.0){\includegraphics[width=3.0cm]{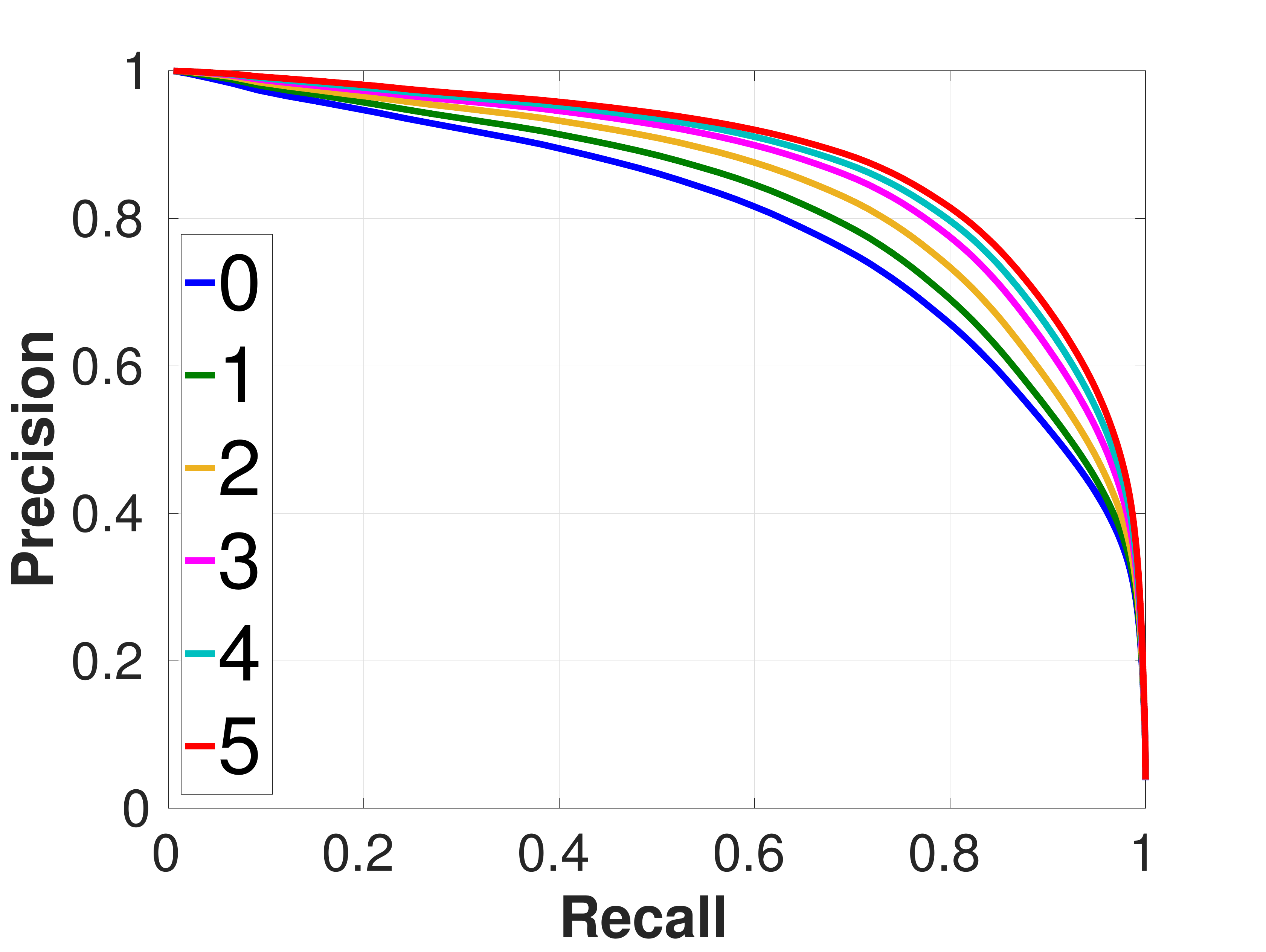}}
        \end{picture}
        \caption{DT LSTM-CNN}
        \label{fig:results_detection_tracking}
    \end{subfigure}
    \vspace{-0.3cm}
    \caption{Fixed-viewpoint liquid tracking and combined detection \& tracking results. Similar to Fig. \ref{fig:results_detection}, the graphs indicate the precision and recall for each of the three networks and the colored lines indicate the variation in the number of slack pixels we allowed for prediction.}
    \label{fig:results_tracking}
    \vspace{-0.5cm}
\end{figure}

For tracking, we evaluated the performance of the networks on locating both visible and invisible liquid, given segmented input (i.e., each pixel classified as liquid, cup, bowl, or background). Because the viewpoint was fixed level with the bowl, the only visible liquid the network was given was liquid as it passed from cup to bowl. \Cref{fig:results_tracking_cnn,fig:results_tracking_wcnn,fig:results_tracking_lstm} show the performance of each of the three networks. As expected, the LSTM-CNN has the best performance. Interestingly, the multi-frame CNN performs better than expected, given that it only sees approximately 1 second's worth of data and has no memory capability.

\vspace{-0.3cm}
\subsection{Combined Detection \& Tracking}
\vspace{-0.3cm}

Fig. \ref{fig:results_detection_tracking} shows the results of combined detection and tracking for the LSTM-CNN. Given a raw color image, the network predicted where both the visible and occluded liquid was. Comparing this to the rest of Fig. \ref{fig:results_tracking}, it is clear that the network was able to do quite well, despite using raw, unstructured input, unlike the other networks in that figure. This strongly suggests that LSTM-CNNs are best suited not only for detecting liquids, but also tracking them.

\vspace{-0.3cm}
\subsection{Preliminary Real Robot Results}
\label{sec:real_results}
\vspace{-0.3cm}

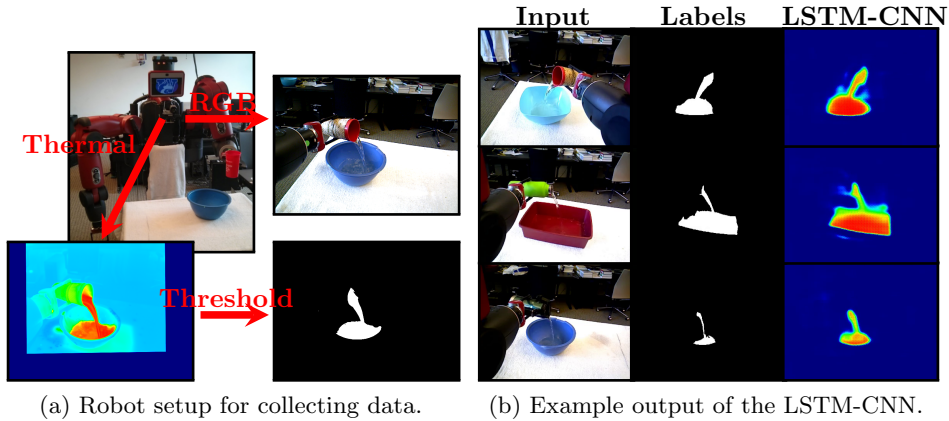
\begin{figure}
    \vspace{-0.3cm}
    \centering
    \setlength{\fboxsep}{0pt}
    \setlength{\fboxrule}{1pt}
    \setlength{\unitlength}{1.0cm}
    \begin{subfigure}{5.5cm}
        \begin{picture}(5.5,4.7)
            \put(0.6,1.7){\fbox{\includegraphics[width=2.4cm]{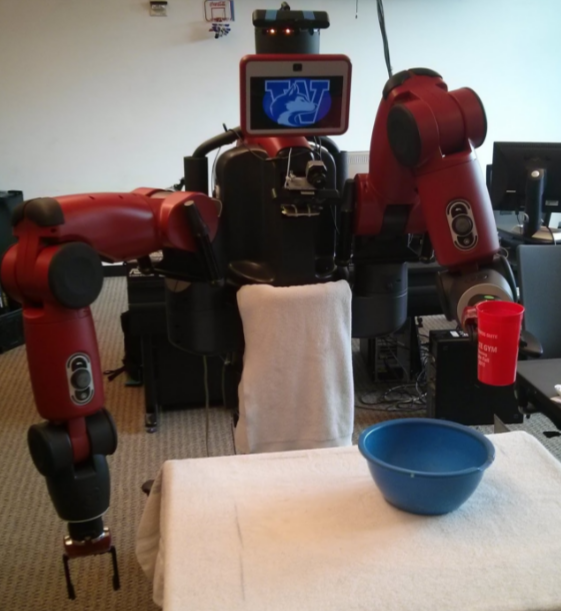}}}
            \put(3.3,2.2){\fbox{\includegraphics[width=2.4cm]{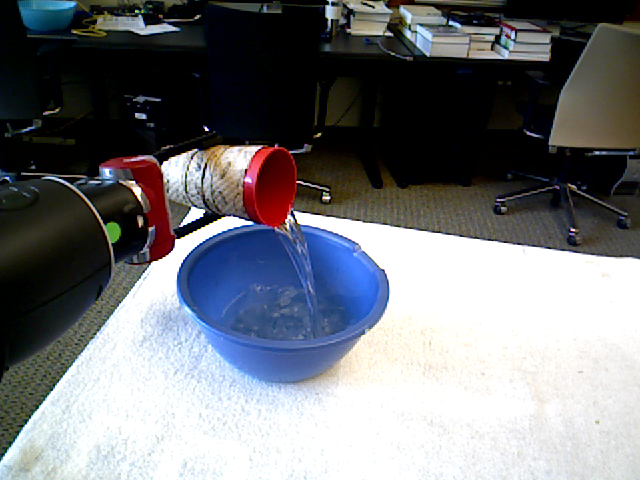}}}
            \put(-0.2,0.0){\fbox{\includegraphics[width=2.4cm]{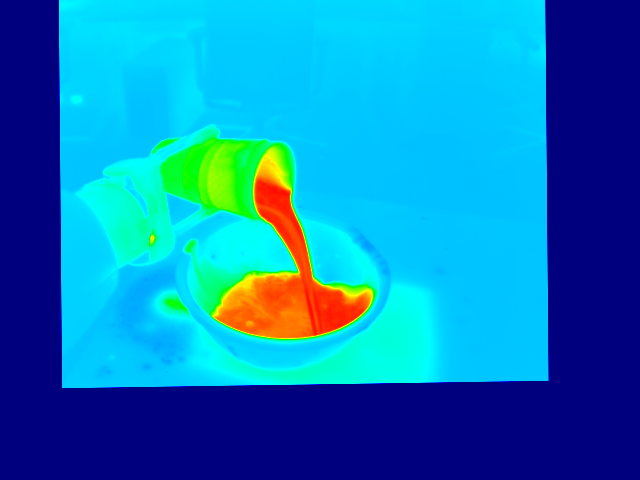}}}
            \put(3.3,0.0){\fbox{\includegraphics[width=2.4cm]{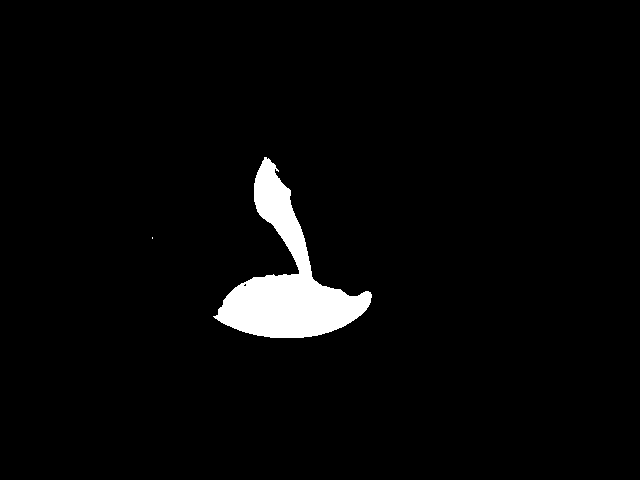}}}
            \put(1.0,1.8){\tikz\draw[red,line width=0.1cm,->,>=stealth] (0.8cm,1.6cm) -- (0,0);}
            \put(0.0,3.0){\color{red}\bf Thermal}
            \put(2.1,3.4){\tikz\draw[red,line width=0.1cm,->,>=stealth] (0.0cm,0.0cm) -- (1.1cm,0.0cm);}
            \put(2.2,3.55){\color{red}\bf RGB}
            \put(2.3,0.8){\tikz\draw[red,line width=0.1cm,->,>=stealth] (0.0cm,0.0cm) -- (0.9cm,0.0cm);}
            \put(1.8,0.95){\color{red}\bf Threshold}
        \end{picture}
        \caption{Robot setup for collecting data.}
        \label{fig:real_setup}
    \end{subfigure}\hspace{0.5cm}%
    \begin{subfigure}{6.0cm}
        \begin{picture}(6.0,4.7)
            \put(0.0,0.0){\fbox{\includegraphics[width=2.0cm]{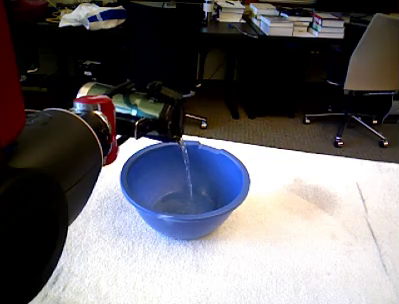}}}
            \put(2.0,0.0){\fbox{\includegraphics[width=2.0cm]{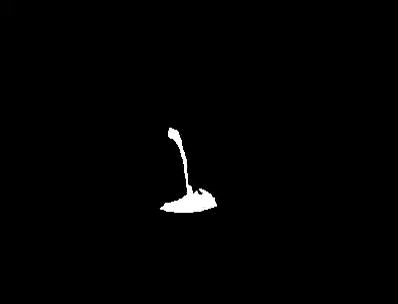}}}
            \put(4.0,0.0){\fbox{\includegraphics[width=2.0cm]{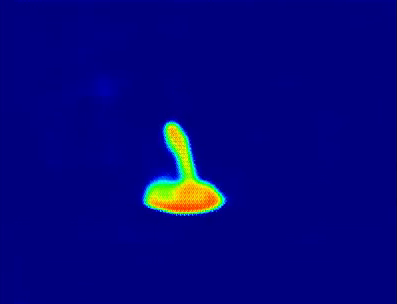}}}
            
            \put(0.0,1.55){\fbox{\includegraphics[width=2.0cm]{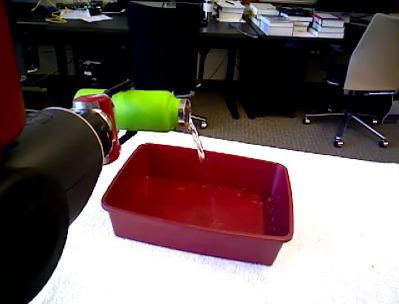}}}
            \put(2.0,1.55){\fbox{\includegraphics[width=2.0cm]{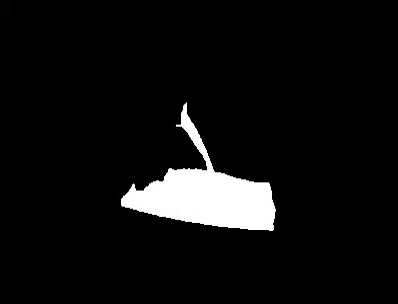}}}
            \put(4.0,1.55){\fbox{\includegraphics[width=2.0cm]{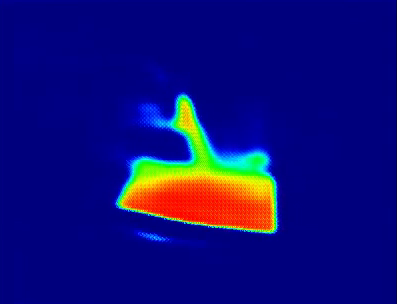}}}

            \put(0.0,3.1){\fbox{\includegraphics[width=2.0cm]{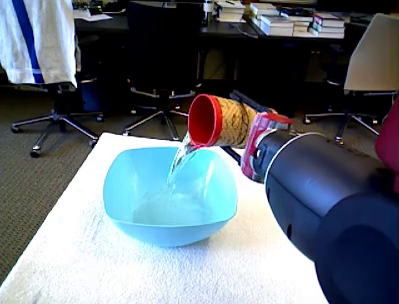}}}
            \put(2.0,3.1){\fbox{\includegraphics[width=2.0cm]{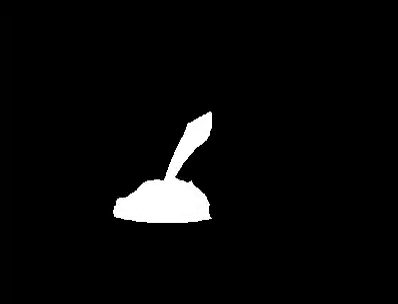}}}
            \put(4.0,3.1){\fbox{\includegraphics[width=2.0cm]{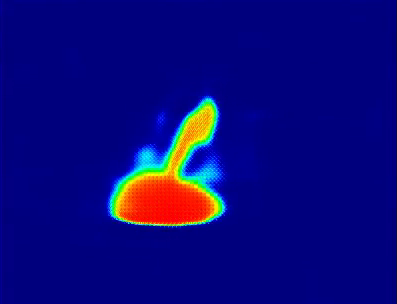}}}
            
            \put(0.5,4.7){{\bf Input}}
            \put(2.4,4.7){{\bf Labels}}
            \put(4.0,4.7){{\bf LSTM-CNN}}
        \end{picture}
        \caption{Example output of the LSTM-CNN.}
        \label{fig:real_results}
    \end{subfigure}
    \caption{The left figure shows our robot, it's attached thermal and RGB cameras, and example output of each camera. The right figure shows the output of a LSTM-CNN trained on data collected on the real robot. The Input column is the input to the network, the Labels column is the ground truth labeling of each pixel as liquid or not liquid, and the LSTM-CNN column shows a heatmap of the prediction of the network for each of the input frames.}
    \label{fig:real_data_collection}
    \vspace{-0.3cm}
\end{figure}

Fig. \ref{fig:real_results} shows qualitative results of the LSTM-CNN trained on a small dataset collected on a real robot in our lab\footnote{Full video of results at \url{https://youtu.be/4pbjSqg5zfQ}}.
We used a thermal infrared camera calibrated to our RGB camera in combination with heated water to acquire ground truth labeling for data collected using a real robot.
The advantage of this method is that heated water appears identical to room temperature water on a standard color camera, but is easily distinguishable on a thermal camera.
This allows us to label the ``hot'' pixels as liquid and all other pixels as not liquid.
Fig. \ref{fig:real_setup} shows our robot setup with the thermal and RGB cameras. 
It is clear from Fig. \ref{fig:real_results} that our methods, to at least a limited degree, apply to real world data and not just data generated by a liquid simulator.

\vspace{-0.5cm}
\section{Discussion \& Conclusion}
\label{sec:conclusion}
\vspace{-0.3cm}

The results in Section \ref{sec:results} show that it is possible for deep learning to detect and track liquids in a scene, both independently and combined, and also over a wide variation in viewpoints.
Unlike prior work on image segmentation, these results clearly show that single images are not sufficient to reliably perceive liquids. 
Intuitively, this makes sense, as a transparent liquid can only be perceived through its refractions, reflections, and specularities, which vary significantly from frame to frame, thus necessitating aggregating information over multiple frames. 
We also found that LSTM-based CNNs are best suited to not only aggregate this information, but also to track the liquid as it moves between containers. 
LSTMs work best, due to not only their ability to perform short term data integration (just like the MF-CNN), but also to remember states, which is crucial for tracking the presence of liquids even when they're invisible.

From the results shown in Fig. \ref{fig:results} and in the video\footnote{Video of the full sequences at \url{\youtubeurl}}, it is clear that the LSTM CNN can at least roughly detect and track liquids.
Nevertheless, unlike the task of image segmentation, our ultimate goal is not to perfectly estimate the potential location of liquids, but to perceive and reason about the liquid such that it is possible to manipulate it using raw sensory data.  
For this, a rough sense of where the liquid is in a scene and how it is moving might suffice.  
Neural networks, then, have the potential to be a key component for enabling robots to handle liquids using robust, closed-loop controllers.

\vspace{-0.5cm}
\section{Future Work}
\label{sec:future_work}
\vspace{-0.5cm}

We are currently on expanding the real robot results from Section \ref{sec:real_results}.
As stated in Section \ref{sec:methodology}, it can be difficult to get the ground truth pixel labels for real data, which is why we chose to use a realistic liquid simulator in this paper. 
However, our method of combing a thermal camera with heated water to get the ground truth makes it feasible to apply the techniques in this paper to data collected on a real robot.
For future work we plan to collect more data on the real robot using this technique and do a thorough analysis of the results.

Another avenue of future work we are currently pursuing is extending these techniques to control problems. 
The results here clearly show that deep neural networks can effectively be used to detect and, to some extent at least, reason about liquids.
The next logical step is to utilize neural networks to manipulate liquids via a robot.
One potential algorithm to accomplish this is Guided Policy Search (GPS) \cite{levine2013}, which learns a control policy for a task from raw sensory data.
The advantage of an algorithm like GPS is that it works well on high-dimensional sensory input where collecting large amounts of data may be infeasible (as is often the case on a real robotic system).
In future work we plan to apply a similar algorithm to the problem of robotic liquid control from raw sensory data.

\vspace{-0.5cm}
\section{Acknowledgments}
\vspace{-0.3cm}

This work was funded in part by the National Science Foundation under contract number NSF-NRI-1525251 and by the Intel Science and Technology Center for Pervasive Computing (ISTC-PC).

\bibliographystyle{splncs}
\vspace{-0.5cm}
{\footnotesize 
\bibliography{iser2016}}

\end{document}